\pdfoutput=1

\documentclass[11pt]{article}
\newcommand{\benchname}{\textsc{Pralekha}}
\newcommand{\dacfull}{\textsc{Document Alignment Coefficient (DAC)}}
\newcommand{\dac}{\textsc{Document Alignment Coefficient}}

 \usepackage[final]{acl}

\usepackage{times}
\usepackage{latexsym}
\usepackage{float}
\usepackage{amsmath}
\usepackage{multirow}
\usepackage{graphicx}
\usepackage{booktabs}
\usepackage{algorithm}
\usepackage{algorithm}
\usepackage{colortbl}
\usepackage{algpseudocode}
\usepackage{amssymb}
\usepackage{microtype}
\usepackage{inconsolata}
\usepackage{hyperref}
\usepackage{fontawesome5}
\usepackage{makecell}

\usepackage{xcolor}
\definecolor{darkpink}{RGB}{204, 51, 102}
\definecolor{darkgreen}{RGB}{0, 128, 0}
\definecolor{darkred}{RGB}{160,0,0}
\definecolor{llamaBlue}{RGB}{0, 4, 255}
\definecolor{sarvamGreen}{RGB}{0, 255, 51}
\definecolor{darkpastelblue}{HTML}{3B76AF}
\definecolor{darkpastelorange}{HTML}{E07A3E}

\usepackage[T1]{fontenc}
\usepackage[utf8]{inputenc}
\title{\benchname{}: Cross-Lingual Document Alignment for Indic Languages}

\author{
  \textbf{Sanjay Suryanarayanan}\textsuperscript{1} \hspace{5pt}
  \textbf{Haiyue Song}\textsuperscript{3} \\ \hspace{5pt}
  \textbf{Mohammed Safi Ur Rahman Khan}\textsuperscript{1,4}
  \hspace{5pt}
  \textbf{Anoop Kunchukuttan}\textsuperscript{1,2} \hspace{5pt}
  \textbf{Raj Dabre}\textsuperscript{1,3,4,5}\thanks{Corresponding author: \href{mailto:raj.dabre@cse.iitm.ac.in}{\texttt{raj.dabre@cse.iitm.ac.in}}} \\
  \textsuperscript{1}Nilekani Centre at AI4Bharat \hspace{5pt}
  \textsuperscript{2}Microsoft  \\
  \textsuperscript{3}National Institute of Information and Communications Technology, Japan \\
  \textsuperscript{4}Indian Institute of Technology, Madras \hspace{5pt}
  \textsuperscript{5}Indian Institute of Technology, Bombay \\
  \href{https://huggingface.co/datasets/ai4bharat/Pralekha}{\raisebox{-0.3em}{\includegraphics[height=1em]{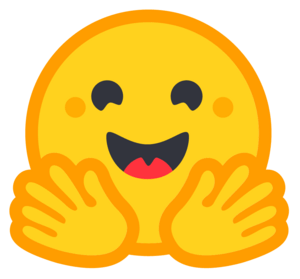}} \hspace{0.1em}\footnotesize huggingface.co/Pralekha} \hspace{0.2em}
  \href{https://github.com/AI4Bharat/Pralekha}{\raisebox{-0.3em}{\includegraphics[height=1em]{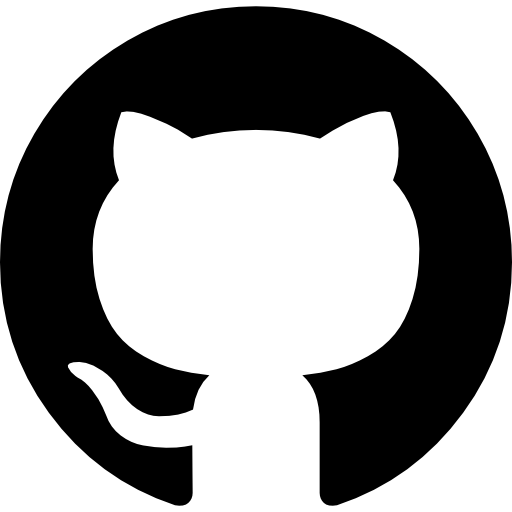}} \hspace{0.1em}\footnotesize github.com/Pralekha} 
}

\begin{document}
\maketitle

\begin{abstract} 
Mining parallel document pairs for document-level machine translation (MT) remains challenging due to the limitations of existing Cross-Lingual Document Alignment (CLDA) techniques. Existing methods often rely on metadata such as URLs, which are scarce, or on pooled document representations that fail to capture fine-grained alignment cues. Moreover, the limited context window of sentence embedding models hinders their ability to represent document-level context, while sentence-based alignment introduces a combinatorially large search space, leading to high computational cost. To address these challenges for Indic languages, we introduce \benchname{}\footnote{\benchname{} means \textit{document} in Sanskrit, an ancient Indo-Aryan language.}, a benchmark containing over 3 million aligned document pairs across 11 Indic languages and English, which includes 1.5 million English–Indic pairs. Furthermore, we propose Document Alignment Coefficient (DAC), a novel metric for fine-grained document alignment. Unlike pooling-based methods, DAC aligns documents by matching smaller chunks and computes similarity as the ratio of aligned chunks to the average number of chunks in a pair. Intrinsic evaluation shows that our \emph{chunk-based method is 2–3× faster while maintaining competitive performance}, and that \emph{DAC achieves substantial gains over pooling-based baselines}. Extrinsic evaluation further demonstrates that document-level MT models trained on DAC-aligned pairs consistently outperform those using baseline alignment methods. \emph{These results highlight DAC’s effectiveness for parallel document mining.} The dataset and evaluation framework are publicly available to support further research. 
\end{abstract}

\begin{figure}[!ht]
    \centering
    \includegraphics[width=0.9\linewidth]{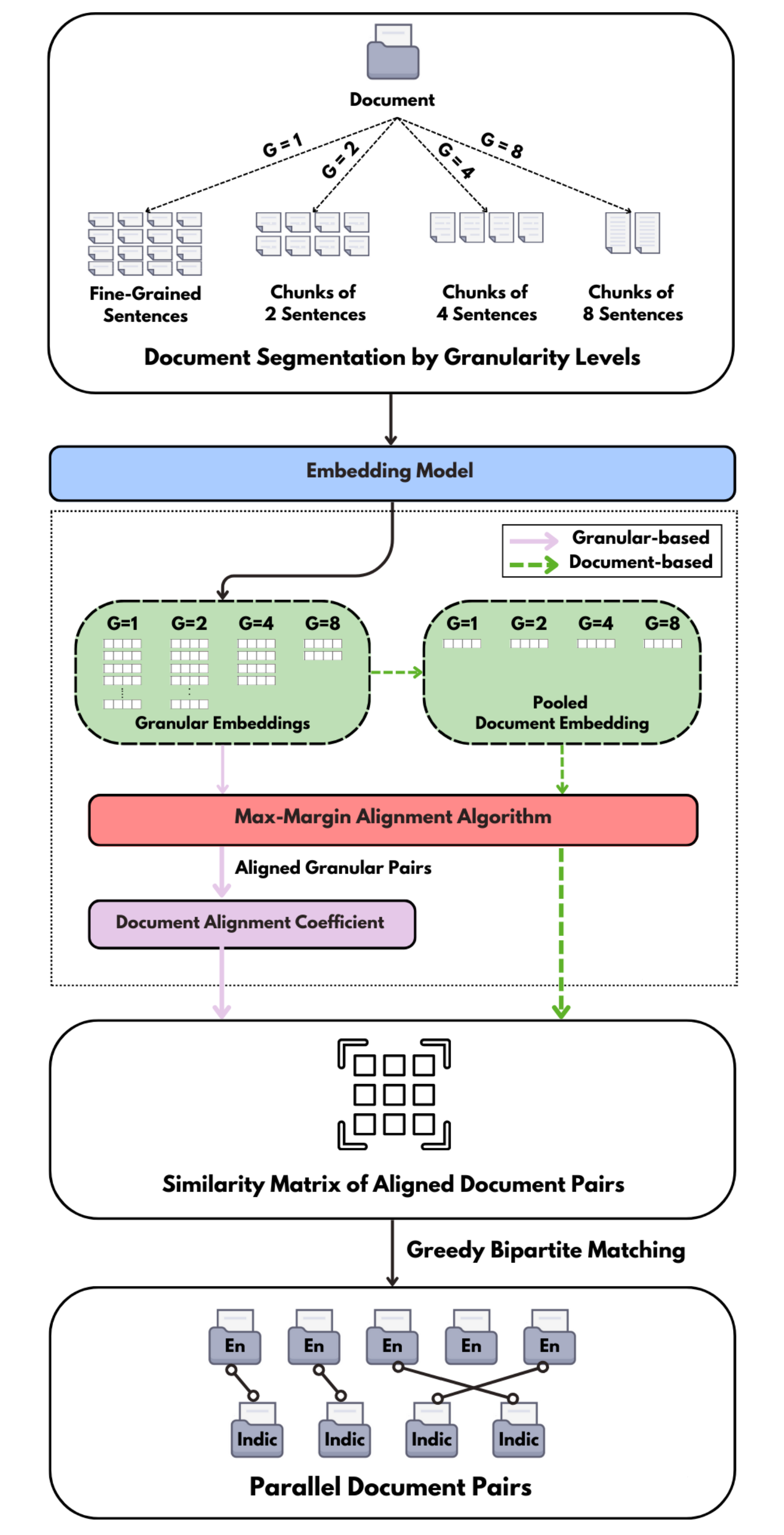}
    \caption{An Overview of the Evaluation Framework for Cross-Lingual Document Alignment. The \textcolor{darkpink}{\textbf{pink path}} illustrates the proposed approach leveraging \dacfull{}, while the \textcolor{darkgreen}{\textbf{green path}} represents pooling-based baseline methods.}
    \label{fig:CLDA}
\end{figure}

\section{Introduction}
\label{sec:introduction}

The advent of Large Language Models (LLMs) has significantly improved the ability to process long-context information, enabling advancements in document-level tasks such as machine translation (MT). Document-level MT extends beyond sentence-level translation by capturing discourse-level dependencies, coherence, and contextual information across entire documents, leading to more accurate and fluent translations. While there are efforts~\cite{ramesh-etal-2022-samanantar, siripragada-etal-2020-multilingual} that provide sentence-level parallel data, large-scale document-level parallel corpora remain scarce~\cite{sannigrahi-etal-2023-best, pal-etal-2024-document}.

To address this gap, parallel document mining is essential for building high-quality training corpora for document-level MT. However, this progress is hindered by two key challenges: \emph{(a) the lack of large-scale document alignment evaluation benchmarks and (b) the limitations of current Cross-Lingual Document Alignment (CLDA) techniques.} Reliable evaluation benchmarks are crucial because the effectiveness of an alignment method cannot be improved without accurate performance measurements. As for the limitations of CLDA, most existing embedding models were developed for sentence-level parallel corpus mining~\cite{schwenk2020ccmatrixminingbillionshighquality, schwenk-etal-2021-wikimatrix, banon-etal-2020-paracrawl}. As a result, they have limited context windows and can represent only parts of a document \cite{zhu2024longembedextendingembeddingmodels}. Document-level embeddings are therefore often obtained by truncating documents or pooling embeddings of independently encoded chunks, both of which can be lossy~\cite{sannigrahi-etal-2023-best}. These approaches overlook chunk-level cues, which reduces their ability to capture complete document semantics. Even without pooling, the reliance of existing alignment techniques on sentence-level operations leads to combinatorial growth in the search space, making large-scale document alignment computationally expensive.

In this paper, we focus on Indic languages, which have abundant sentence-level parallel corpora and benchmarks ~\cite{ramesh-etal-2022-samanantar, siripragada-etal-2020-multilingual, gala2023indictrans2highqualityaccessiblemachine} but remain low-resource for CLDA and Document-level MT. We first introduce \benchname{}, a large-scale parallel document dataset for evaluating CLDA techniques across 11 Indic languages and English. \benchname{} comprises over 3 million document pairs between 11 Indic languages and English, of which 1.5 million are English--Indic pairs. These are high-quality, human-verified aligned documents sourced from reliable domains such as the Indian Press Information Bureau (PIB) and the Mann Ki Baat (MKB) radio program. It serves both as a benchmark for assessing CLDA and as a domain-specific parallel corpus for training document-level MT models for Indic languages. Alongside \benchname{}, we propose \dacfull{}, a novel metric for CLDA at the fine-granular level. Unlike existing approaches that rely on pooled document-level embeddings, we focus on smaller granular units, called chunks. We first embed and align these chunks, then compute the DAC of a document pair as the ratio of aligned chunks to the average number of chunks in the pair.

In our experiments using \benchname{} under realistic noisy document mining settings, we evaluate the proposed DAC approach against existing pooling-based baselines. \emph{DAC achieves average precision, recall, and F1 scores of 0.8932, 0.6312, and 0.7372, respectively, showing improvements of 15–20\% in precision and 5–10\% in F1 score over the baselines.} We further benchmark DAC on the CCAligned~\cite{el-kishky-etal-2020-ccaligned} dataset, where it attains average precision, recall, and F1 scores of 0.8110, 0.6511, and 0.7203, respectively, outperforming the baselines across all metrics. Additionally, \emph{our chunk-based method demonstrates a 2–3× speedup over sentence-based alignment techniques while maintaining competitive performance}. Having established DAC’s intrinsic effectiveness, we conduct an extrinsic evaluation by training document-level MT models on mined documents using open-source LLMs. The results show that \emph{MT models trained on DAC-aligned pairs consistently outperform those trained with baseline alignment methods}. The \benchname{} dataset and CLDA evaluation framework is publicly available to support further research in this area.
\section{Related Work}
\label{sec:related-work}

\noindent\textbf{Cross-Lingual Sentence Alignment.}  
Early approaches to sentence-level alignment, such as Europarl~\cite{koehn-2005-europarl}, relied on manually curated metadata~\cite{abdul-rauf-schwenk-2009-improving, do-etal-2009-mining}. Recent advancements in multilingual embeddings have significantly improved large-scale mining of parallel data from unstructured sources. Methods such as LaBSE~\cite{feng-etal-2022-language}, LASER~\cite{schwenk-douze-2017-learning}, SONAR~\cite{duquenne2023sonarsentencelevelmultimodallanguageagnostic}, and hierarchical bilingual retrieval~\cite{guo-etal-2019-hierarchical} leverage multilingual sentence representations to enhance cross-lingual alignment accuracy. These techniques have enabled large-scale efforts such as CCMatrix~\cite{schwenk2020ccmatrixminingbillionshighquality}, WikiMatrix~\cite{schwenk-etal-2021-wikimatrix}, ParaCrawl~\cite{banon-etal-2020-paracrawl}, NLLB~\cite{nllbteam2022languageleftbehindscaling}, and BPCC~\cite{gala2023indictrans2highqualityaccessiblemachine}, which extract and align sentence pairs from massive web crawls. 

\noindent\textbf{Cross-Lingual Document Alignment.}  
For document-level alignment, early approaches relied on metadata rather than content, using features such as URL and page structure similarity~\cite{resnik-smith-2003-web}. While multilingual encoders have improved sentence-level parallel data extraction, document-level alignment remains challenging due to the limited context windows of existing sentence embedding models~\cite{zhu2024longembedextendingembeddingmodels}. In this work, we explore a faster chunk-based method that effectively leverages fine-granular embeddings for document-level alignment.

\noindent\textbf{Indic Language Corpora.}
Despite India’s large population (1.43 billion), its languages remain low-resource, particularly for high-quality document-level data. Significant efforts have been made to build monolingual corpora, such as IndicNLP Suite~\cite{kakwani-etal-2020-indicnlpsuite} and Sangraha~\cite{khan-etal-2024-indicllmsuite}, as well as parallel corpora and benchmarks like IN22 and BPCC~\cite{gala2023indictrans2highqualityaccessiblemachine}. However, no publicly available cross-lingual parallel document datasets exist for Indic languages~\cite{sannigrahi-etal-2023-best}. The only work referencing parallel documents uses them solely for mining sentence pairs, without releasing the document-level data~\cite{siripragada-etal-2020-multilingual}. This underscores the urgent need to construct parallel document corpora for cross-lingual document alignment and document-level machine translation, a gap that our work aims to fill.

\section{\benchname{}}

\begin{figure}[h]
    \centering
    \includegraphics[width=\linewidth]{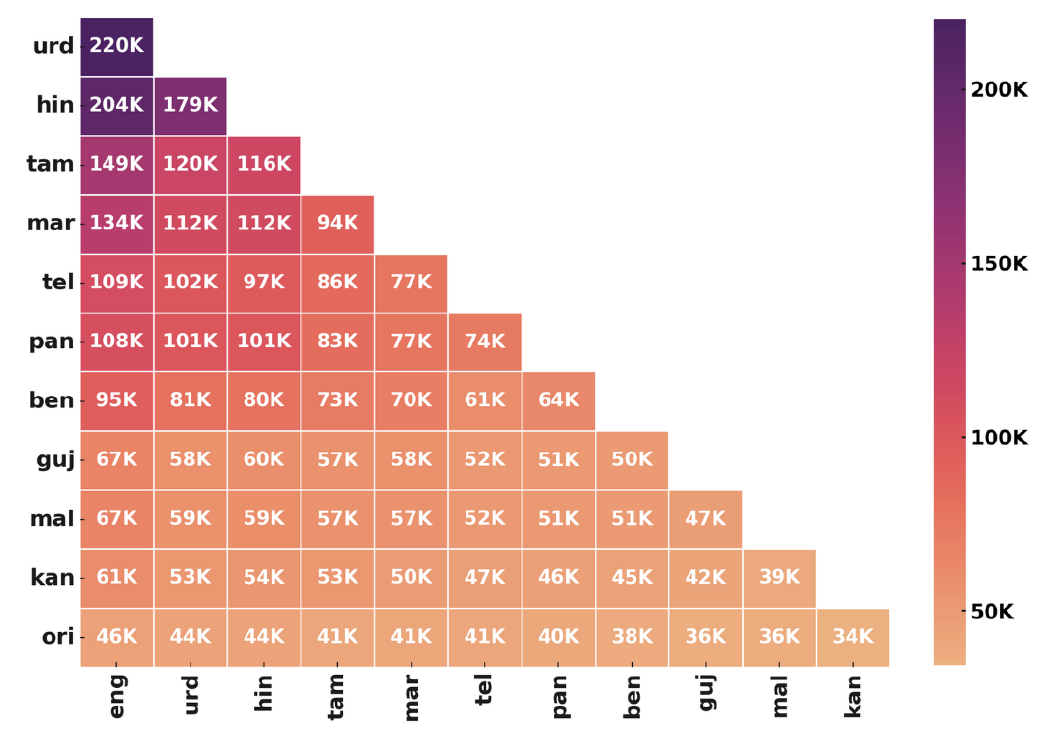}
    \caption{Heat-map of \textit{alignable} document pairs for each language pair in \benchname{}. Darker cells indicate higher alignment counts.}
    \label{fig:aligned-pairs_heatmap}
\end{figure}

\label{sec:benchmark}
We present \benchname{} a large-scale parallel document dataset for evaluating Cross-Lingual Document Alignment (CLDA) techniques.
\benchname{} comprises over 3 million document pairs between 12 languages - Bengali (ben), Gujarati (guj), Hindi (hin), Kannada (kan), Malayalam (mal), Marathi (mar), Odia (ori), Punjabi (pan), Tamil (tam), Telugu (tel), Urdu (urd) and English (eng) containing a mixture of high, medium and low resource languages. Figure~\ref{fig:aligned-pairs_heatmap} shows a heatmap of the distribution of \textit{alignable} document pairs across language combinations. Among these, 1.5 million document pairs are English--Indic. 

\begin{figure}[!h]
    \centering
    \includegraphics[width=\linewidth]{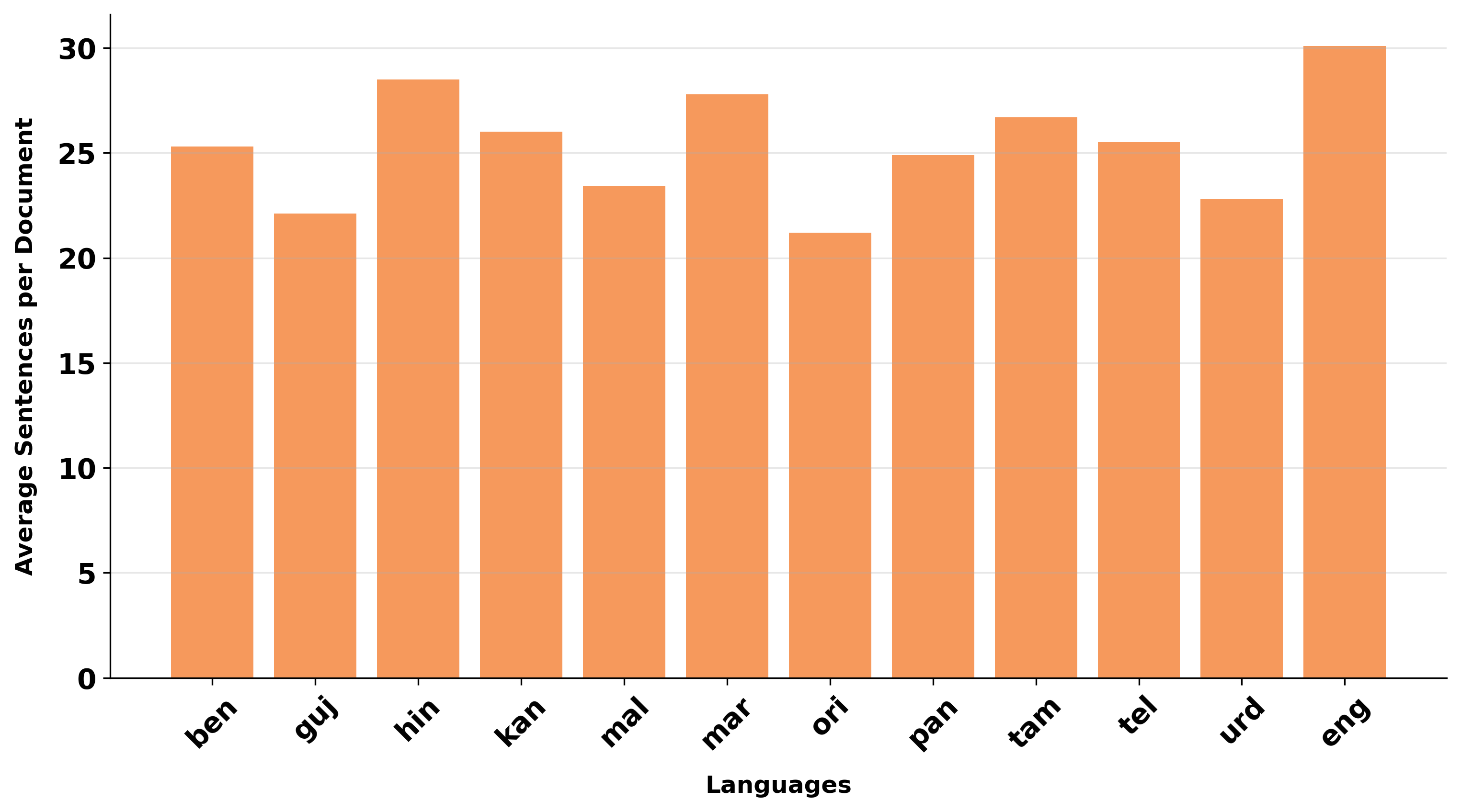}
    \caption{Average document length in \benchname{} for each language, measured in terms of number of sentences per document.}
    \label{fig:sent_count_plot}
\end{figure}
Figure~\ref{fig:sent_count_plot} presents statistics on the average number of sentences per document for each language in \benchname{}. The average sentence count ranges from 20 to 30, indicating that the documents are relatively long and making the benchmark more challenging.

\benchname{} covers two broad domains - News Bulletins and Podcast Scripts, therefore containing a mixture of written as well as spoken form of data. Following the methodology of Sangraha~\cite{khan-etal-2024-indicllmsuite}, we custom-scraped data from the Indian Press Information Bureau (PIB)\footnote{\url{https://pib.gov.in}} website, aligning documents by matching bulletin IDs interlinking bulletins across languages. These documents are manually written and hence are of high quality. For podcast scripts, we employed the approach used by~\citet{khan-etal-2024-indicllmsuite} and~\citet{siripragada-etal-2020-multilingual} to collect transcripts from Mann Ki Baat,\footnote{\url{https://www.pmindia.gov.in/en/mann-ki-baat}} a radio program hosted by the Indian Prime Minister. This program is typically spoken in Hindi, then manually transcribed and translated into various other Indian languages. Sometimes, the website's meta-data incorrectly marks the document's languages, which we fix with a simple combination of IndicLID \cite{madhani-etal-2023-bhasa-abhijnaanam} and script-specific Unicode ranges. 

All the data is human-written/verified. \textit{This dataset serves both as a benchmark for assessing document alignment techniques and as a domain-specific parallel corpus for training document-level MT models in Indic Languages.}
\section{Cross-Lingual Document Alignment}
\label{sec:methodology}

Given two document collections, \( D_1 \) and \( D_2 \), where \( D_1 \) consists of documents written in \texttt{lang1} and \( D_2 \) consists of documents in \texttt{lang2}, the objective of Cross-Lingual Document Alignment (CLDA) is to determine a set of paired documents,  \( P = \{ (d_{1}, d_{2}) \mid d_{1} \in D_1, d_{2} \in D_2 \} \), such that each pair \( (d_{1}, d_{2}) \) exhibits a high degree of semantic similarity, with \( d_1 \) and \( d_2 \) being direct translations of each other. 

We next describe the evaluation framework used to benchmark various CLDA techniques. As illustrated in Figure~\ref{fig:CLDA}, the framework includes our proposed approach, which leverages the \dacfull{}, along with the pooling-based baseline methods. The following subsections discuss each component of the CLDA evaluation framework in detail.

\subsection{Granularity}
\label{subsec:granularity}
The granularity level \( G \) defines the unit of text at which alignment is performed and is a key parameter in our study. We evaluate document alignment at three levels of granularity: \( G = 1 \) (sentence-level, as in \citet{schwenk2020ccmatrixminingbillionshighquality}), \( G = 2, 4, 8 \) (chunk-level, where each chunk consists of 2, 4, or 8 sentences, respectively), and \( G = |D| \) (document-level, as in \citet{el-kishky-etal-2020-ccaligned, el-kishky-guzman-2020-massively}). 

At one extreme, \( G = 1 \) produces too many embeddings to align, resulting in a combinatorially large search space that leads to high computational cost, while offering limited context for semantic comparison. At the other extreme, \( G = |D| \) leads to overly coarse representations and information loss. It also creates challenges in computing a single document-level embedding because embedding models have a limited context window~\cite{zhu2024longembedextendingembeddingmodels}. Both extremes have limitations; therefore, we consider intermediate values of \( G \) that strike a balance between efficiency and information retention. A larger granularity helps preserve local coherence and capture broader document-level information, addressing the shortcomings of traditional sentence-based approaches~\cite{xie2024chunkalignselectsimple}.

\noindent \textbf{Pooling Strategies for Baselines.} We benchmark several pooling strategies used in prior work \cite{el-kishky-etal-2020-ccaligned, el-kishky-guzman-2020-massively} to construct document embeddings (\( G = |D| \)) by aggregating finer-grained unit embeddings.

\noindent 1. \textsc{Mean Pooling (MP)}: Computes a single embedding by averaging all unit embeddings.  

\noindent 2. \textsc{Length Pooling (LP)}: Adjusts unit embeddings based on length, assigning greater importance to longer, more informative segments.  

\noindent 3. \textsc{Inverse Document Frequency} (IDF): It weighs terms according to their rarity, prioritizing informative words while down-weighting common ones to improve alignment precision.  

\noindent 4. \textsc{Length-Inverse Document Frequency (LIDF)}: Combines length-based weighting and IDF, emphasizing longer, content-rich segments while reducing the influence of repetitive or generic terms, providing a balanced approach to alignment.

\subsection{Embedding Model}

To compute multilingual embeddings we consider {\textsc{LaBSE} \hspace{0pt} \raisebox{-0.1\height}{\includegraphics[height=10pt]{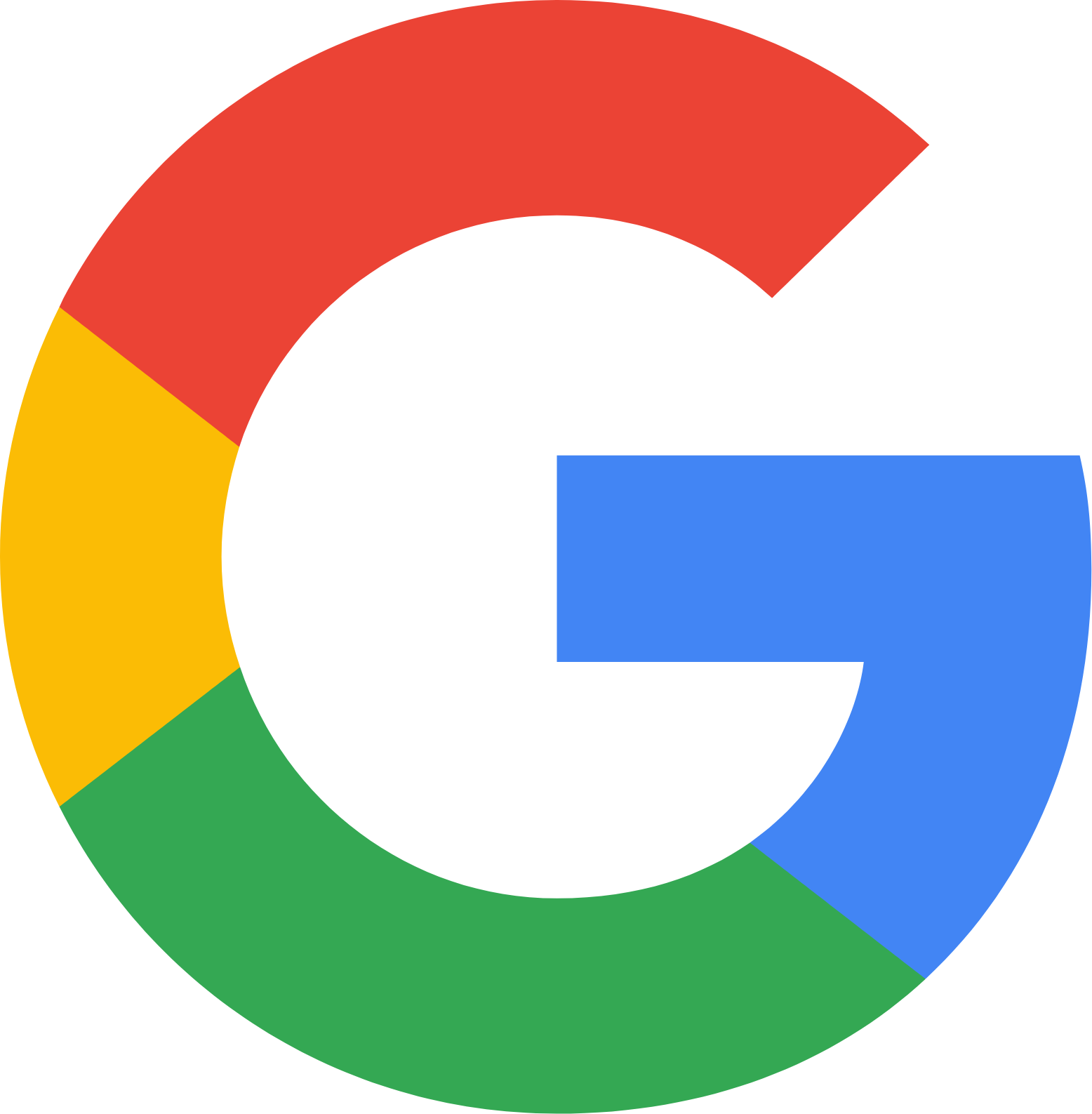}}} \cite{feng-etal-2022-language} and {\textsc{SONAR} \hspace{-1pt} \raisebox{-0.2\height}{\includegraphics[height=12pt]{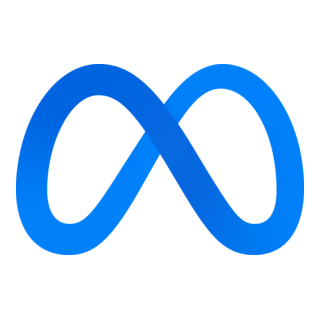}}} \cite{duquenne2023sonarsentencelevelmultimodallanguageagnostic}.  
\textsc{LaBSE} is designed for cross-lingual sentence retrieval, leveraging both parallel and monolingual data to generate language-agnostic embeddings. \textsc{SONAR} extends this approach by training on a more diverse set of languages, leading to improved representation quality for both high-resource and low-resource languages.
\noindent \textbf{Computing Embeddings.} After segmenting each document into units based on the chosen granularity \( G \), we obtain representations for each segment by encoding the entire unit with the embedding model. Because the values of \( G \) considered in our study are relatively small, each segment fits within the context window of the embedding models used.

\subsection{Alignment Algorithm}
After computing unit embeddings at various text granularities from the monolingual corpora in the source and target languages, we apply the Alignment Algorithm~\ref{algo:alignment} to align these embeddings. The algorithm performs margin-based bitext mining in a shared multilingual space to extract parallel units~\cite{schwenk2020ccmatrixminingbillionshighquality}. It adopts the max-margin criterion, which has been shown to improve alignment quality compared to an absolute similarity threshold~\cite{artetxe-schwenk-2019-margin}. To build efficient similarity indices, we use the FAISS library\footnote{\url{https://github.com/facebookresearch/faiss}}. For each unit embedding, we retrieve the top-$k$ nearest neighbors and compute margin scores \( M(x, y) \) to capture similarity in a relative context. Candidate pairs are then ranked by their margin scores, and a greedy bipartite matching strategy is applied to select the highest-scoring pairs, ensuring that each embedding is aligned only once.

\begin{algorithm}[!h]
\caption{Alignment Algorithm} \label{algo:alignment}
\begin{algorithmic}[1]
\small
\State \textbf{Input:} Embeddings $\mathbf{X} = \{x_i\}$ and $\mathbf{Y} = \{y_j\}$, $k$ nearest neighbors
\vspace{5pt}
\State \textbf{Output:} $\mathcal{U} = \arg\max_{\mathcal{U}'} \sum_{(x, y) \in \mathcal{U}'} M(x, y)$
\vspace{5pt}

\State \textbf{FAISS Indexing:} Construct FAISS index for efficient similarity search
\vspace{-7pt}
\[
\texttt{index\_X} \gets \texttt{IndexFlatIP}(\mathbf{X})
\]
\vspace{-17pt}
\[
\texttt{index\_Y} \gets \texttt{IndexFlatIP}(\mathbf{Y})
\]

\State \textbf{Query top-$k$ nearest neighbors:}
\vspace{-7pt}
\[
\texttt{NN}_k(x) = \{ y_1, y_2, \dots, y_k \} \subset \mathbf{Y} \text{ for each } x \in \mathbf{X}
\]
\vspace{-17pt}
\[
\texttt{NN}_k(y) = \{ x_1, x_2, \dots, x_k \} \subset \mathbf{X} \text{ for each } y \in \mathbf{Y}
\]

\State \textbf{Compute Margin Score:}
    \For{each $(x, y)$ in $\mathbf{X} \times \mathbf{Y}$ where $y \in \texttt{NN}_k(x)$ or $x \in \texttt{NN}_k(y)$}
        \begin{small}
        \[
        A_x = \frac{1}{k} \sum_{z \in \texttt{NN}_k(x)} \cos(x, z)
        \]
        \[
        A_y = \frac{1}{k} \sum_{z \in \texttt{NN}_k(y)} \cos(y, z)
        \]
        \end{small}
        \[
        M(x, y) = \frac{\cos(x, y)}{0.5 \times (A_x + A_y)}
        \]
    \EndFor
\vspace{5pt}

\State \textbf{Max Strategy:} Select pairs to maximize overall margin score
\begin{small}
\[
\mathcal{U} = \left\{ (x, y) \mid y \in \texttt{NN}_k(x) \right\} 
\]
\vspace{-17pt}
\[
\cup \left\{ (x, y) \mid x \in \texttt{NN}_k(y) \right\}
\]
\vspace{-5pt}
\[
\mathcal{U}_{\text{sorted}} = \text{Sort}(\mathcal{U}) \text{ by } M(x, y) \text{ in descending order}
\]
\end{small}

\For{each $(x, y)$ in $\mathcal{U}_{\text{sorted}}$}
    \If{both $x$ and $y$ are not in aligned pairs}
        \State Add $(x, y)$ to aligned pairs
    \EndIf
\EndFor

\State \textbf{Return:} $\mathcal{U}$, aligned pairs that maximize the overall margin score

\end{algorithmic}
\end{algorithm}

\emph{For pooled document embeddings at \( G = |D| \), we directly apply the alignment algorithm on these embeddings to obtain aligned document pairs.} We extend this idea by introducing a novel \dacfull{} metric for mining parallel documents, which leverages finer-grained units for document alignment. The motivation behind DAC is to avoid relying solely on document-level embeddings, which are typically formed by pooling unit embeddings and may fail to capture detailed semantic information. Instead, we treat each document as a sequence of chunks, where each chunk is a contiguous set of sentences and serves as the basic unit of alignment. Our hypothesis is that embeddings at finer granularity capture semantic nuances more effectively. After performing alignment at the chunk level, the resulting information is aggregated to identify aligned document pairs. The following subsection provides a detailed description of the DAC approach.

\subsection{\dac{}}
Unlike baseline approaches that operate directly on pooled document embeddings, DAC performs chunk-level alignment first and then derives document pairs using a computed score based on these chunk alignments, thereby providing finer-grained semantic signals that improve alignment precision.

\paragraph{Computing DAC for a Document Pair.} For a given source–target document pair, the DAC is computed as follows:

\begin{equation}
\text{DAC} = \frac{2 \times N_{\text{aligned}}}{N_{\text{src}} + N_{\text{tgt}}}
\label{eq:DAC}
\end{equation}  

Here, \( N_{\text{src}} \) and \( N_{\text{tgt}} \) represent the total number of chunks in the source and target documents, respectively, and \( N_{\text{aligned}} \) denotes the number of aligned chunks between the two documents.

The algorithm first aligns the constituent chunks within each document pair and then uses these fine-grained alignments to compute the overall degree of alignment at the document level. 

\paragraph{DAC Threshold.} The DAC produces a normalized alignment score between 0 and 1, where higher values indicate stronger alignment, and a threshold on this score can be set to control the final alignment quality. A lower DAC threshold allows a balance between quality and quantity, accepting some misalignments while increasing coverage. On the other hand, a higher threshold prioritizes precision, ensuring fewer incorrect alignments at the cost of reduced yield. The threshold can be adjusted based on the specific requirements of the aligned dataset and its intended downstream task. A detailed analysis of various DAC Thresholds is provided in Figure~\ref{fig:dac_curve} and Table~\ref{tab:dac-intrinsic-eval-results} in Appendix~\ref{sec:intrinsic-trends}.

\begin{table*}[!ht]
    \small
    \centering
    \begin{tabular}{lcccccccccccc}
        \toprule
        \textbf{Language} & \textbf{ben} & \textbf{guj} & \textbf{hin} & \textbf{kan} & \textbf{mal} & \textbf{mar} & \textbf{ori} & \textbf{pan} & \textbf{tam} & \textbf{tel} & \textbf{urd} & \textbf{eng} \\
        \midrule
        \textit{\textbf{Alignable}}   & 95K  & 67K  & 204K & 61K  & 67K  & 134K & 46K  & 108K & 149K & 109K & 220K & 298K \\
        \textit{\textbf{Unalignable}} & 47K  & 34K  & 102K & 31K  & 34K  & 67K  & 23K  & 54K  & 75K  & 55K  & 110K & 149K \\
        \cmidrule(lr){1-13} 
        \textbf{Total}     & 142K & 101K & 306K & 92K  & 101K & 201K & 69K  & 162K & 224K & 164K & 330K & 447K \\
        \bottomrule
    \end{tabular}
    \caption{Per language statistics of unique documents divided into two groups: \textit{alignable} and \textit{unalignable}. \textit{Alignable} documents come from the English-Indic part of \benchname{} and the \textit{Unalignable} documents come from \textsc{Sangraha Unverified}, representing noise for a realistic evaluation setting.}
    \label{tab:pralekha_stats}
\end{table*}
\section{Experiments}
\label{sec:experiments}
In this section, we present the experimental setup and describe the two types of evaluations conducted on \benchname{}: intrinsic evaluation, which measures the effectiveness of document alignment techniques, and extrinsic evaluation, which assesses their impact on downstream tasks. \textit{Unless stated otherwise, all references to \benchname{} hereafter refer to its English–Indic subset.}

\subsection{Intrinsic Evaluation}  \label{sec:intrinsic-eval}
For intrinsic evaluation, we assess how well DAC performs compared to existing pooling-based baselines for document alignment.

\paragraph{Data Setting :} \benchname{} contains only parallel (\textit{alignable}) documents, so aligning only these would not reflect a realistic and challenging scenario. To simulate real-world multilingual corpora setting, we sample random (\textit{unalignable}) documents from \textsc{Sangraha Unverified}~\cite{khan-etal-2024-indicllmsuite} that cannot be aligned with those in \benchname{} and inject them as noise into the dataset. We select a number of unalignable documents equal to 50\% of the documents in \benchname{}, resulting in a 1:2 ratio of \textit{unalignable} to \textit{alignable} documents during alignment. Table~\ref{tab:pralekha_stats} reports the counts of unique \textit{alignable} and \textit{unalignable} documents for each language used in our experiments. Additionally, we compare the best-performing baseline with our proposed DAC approach on the CCAligned dataset~\cite{el-kishky-etal-2020-ccaligned}, using a similar data setting.

\paragraph{Alignment Settings :}  
We evaluate alignment at four granularities: sentence-level (\(G = 1\)), chunk-level (\(G = 2, 4, 8\)), and document-level (\(G = |D|\)). For \(G = |D|\), we benchmark all pooling-based baselines described in Section~\ref{subsec:granularity}. Unit embeddings are obtained using {\textsc{LaBSE} \raisebox{-0.1\height}{\includegraphics[height=10pt]{images/google.png}}}~\cite{feng-etal-2022-language} and {\textsc{SONAR} \raisebox{-0.2\height}{\includegraphics[height=12pt]{images/meta_logo.png}}}~\cite{duquenne2023sonarsentencelevelmultimodallanguageagnostic}. We build FAISS \texttt{IndexFlatIP}\footnote{\url{https://github.com/facebookresearch/faiss/wiki/The-index-factory}} indices and retrieve the top-\(k\) nearest neighbors for each embedding with \(k = 16\). A DAC threshold of 0.1 is used to balance precision and recall, and a detailed threshold analysis is provided in Figure~\ref{fig:dac_curve} and Table~\ref{tab:dac-intrinsic-eval-results} in the Appendix.

\paragraph{Evaluation Metrics :}  To assess the intrinsic performance of various document alignment methods, we evaluate three key metrics: precision, recall, and F1 score. While a high F1 score is generally desirable as it balances precision and recall, the ideal metric depends on the specific downstream task. We further measure computational time to analyze efficiency trade-offs across different granularities.

\begin{table*}[!ht]
\footnotesize
\centering
\begin{tabular}{llcccccccc}
\toprule
\multirow{2}{*}{\textbf{Metrics}} & \multirow{2}{*}{\textbf{Methods}} 
  & \multicolumn{4}{c}{\textbf{\textsc{LaBSE}} \hspace{3pt} \raisebox{-0.1\height}{\includegraphics[height=10pt]{images/google.png}}}
  & \multicolumn{4}{c}{\textbf{\textsc{SONAR}} \hspace{3pt} \raisebox{-0.2\height}{\includegraphics[height=12pt]{images/meta_logo.png}}} \\
\cmidrule(lr){3-6} \cmidrule(lr){7-10}
  &   & \textbf{G = 1} & \textbf{G = 2} & \textbf{G = 4} & \textbf{G = 8} 
      & \textbf{G = 1} & \textbf{G = 2} & \textbf{G = 4} & \textbf{G = 8} \\
\midrule
\multirow{5}{*}{Precision} 
  & MP    & 0.7585 & 0.7432 & 0.7232 & 0.7018 & 0.7803 & 0.7643 & 0.7541 & 0.7632 \\
  & LP    & 0.7820 & 0.7709 & 0.7614 & 0.7464 & 0.7943 & 0.7843 & 0.7891 & 0.8111 \\
  & IDF   & 0.7610 & 0.7463 & 0.7290 & 0.7052 & 0.7821 & 0.7669 & 0.7596 & 0.7661 \\
  & LIDF  & 0.7831 & 0.7716 & 0.7619 & 0.7466 & 0.7950 & 0.7849 & 0.7894 & 0.8112 \\
  \cmidrule(lr){2-10}
  & DAC  & \textbf{0.9152} & \textbf{0.9007} & \textbf{0.8936} & \textbf{0.8912} & \textbf{0.9106} & \textbf{0.8870} & \textbf{0.8742} & \textbf{0.8732} \\
\midrule
\multirow{5}{*}{Recall} 
  & MP    & 0.6310 & 0.6057 & 0.5653 & 0.5599 & 0.6613 & 0.6349 & 0.5696 & 0.5414 \\
  & LP    & 0.6824 & 0.6807 & 0.7019 & 0.6959 & 0.6924 & 0.6834 & 0.6688 & 0.6630 \\
  & IDF   & 0.6437 & 0.6192 & 0.5962 & 0.5738 & 0.6702 & 0.6471 & 0.5991 & 0.5544 \\
  & LIDF  & \textbf{0.6908} & \textbf{0.6844} & \textbf{0.7030} & \textbf{0.6971} & \textbf{0.6953} & \textbf{0.6864} & \textbf{0.6700} & \textbf{0.6632} \\
  \cmidrule(lr){2-10}
  & DAC & 0.6588 & 0.6411 & 0.6471 & 0.6407 & 0.6484 & 0.6108 & 0.6005 & 0.6020 \\
\midrule
\multirow{5}{*}{F1 Score} 
  & MP    & 0.6856 & 0.6639 & 0.6291 & 0.6157 & 0.7133 & 0.6904 & 0.6410 & 0.6237 \\
  & LP    & 0.7252 & 0.7199 & 0.7273 & 0.7170 & 0.7372 & 0.7276 & 0.7213 & 0.7266 \\
  & IDF   & 0.6940 & 0.6733 & 0.6519 & 0.6266 & 0.7193 & 0.6989 & 0.6647 & 0.6353 \\
  & LIDF  & 0.7303 & 0.7223 & 0.7281 & 0.7178 & 0.7392 & \textbf{0.7297} & \textbf{0.7221} & \textbf{0.7268} \\
  \cmidrule(lr){2-10}
  & DAC  & \textbf{0.7635} & \textbf{0.7463} & \textbf{0.7480} & \textbf{0.7428} & \textbf{0.7550} & 0.7210 & 0.7098 & 0.7109 \\
\bottomrule
\end{tabular}
\caption{Average Precision, Recall, and F1 scores from the Intrinsic Evaluation of CLDA techniques on \benchname{}, computed over 11 Indic languages. Results are shown for \textsc{LaBSE} and \textsc{SONAR} embeddings at granularities \(G = 1, 2, 4, 8\). We compare pooling-based baselines (\textsc{MP}, \textsc{LP}, \textsc{IDF}, \textsc{LIDF}) with our proposed \textsc{DAC} approach. \textbf{Bold} values indicate the best score for each granularity. Appendix~\ref{sec:intrinsic-trends} provides further analysis on the intrinsic trends.}
\label{tab:intrinsic-eval-results}
\end{table*}

\subsection{Extrinsic Evaluation}  
\label{sec:extrinsic-eval}

To assess the extrinsic performance of document alignment techniques, we evaluate their effectiveness in document-level MT tasks.

\paragraph{Data Setting :} We focus on translation between English and eight Indic languages: Bengali, Gujarati, Hindi, Kannada, Malayalam, Marathi, Odia, and Tamil. From \benchname{}, we randomly sample 1,000 document pairs each for the development and test sets. After removing these pairs, we perform document alignment on the remaining data as described in Section~\ref{sec:intrinsic-eval}. We obtain two sets of aligned documents: one using the best-performing baseline and the other using our proposed DAC approach. From each set, we sample 10,000 aligned pairs to train two separate document-level MT models using open-source LLMs. Extrinsic evaluation is then conducted on the sampled test set.

\paragraph{Implementation and Training :} 
We fine-tune two open-source LLMs: \textsc{Llama-3.2-1B}\footnote{\url{huggingface.co/meta-llama/Llama-3.2-1B}} \hspace{-2pt} \raisebox{-0.1\height}{\includegraphics[height=10pt]{images/meta_logo.png}} and \textsc{Sarvam-1}\footnote{\url{huggingface.co/sarvamai/sarvam-1}} \hspace{-2pt} \raisebox{-0.2\height}{\includegraphics[height=12pt]{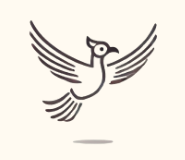}}. \textsc{LLaMA-3.2-1B} is a 1 billion parameter multilingual model released by Meta AI, while \textsc{Sarvam-1} is a 2 billion parameter model developed by Sarvam AI, optimized for Indic languages. We used \textit{open-instruct}\footnote{\url{https://github.com/allenai/open-instruct}} for \textit{full fine-tuning} using the prompt-completion template. We used default hyperparameters recommended in \textit{open-instruct}, most notably: learning rates of \(5 \times 10^{-5}\), maximum context (prompt+completion) lengths of 4096, Adam optimizer, and no weight decay. Training was performed on NVIDIA H100 GPUs with mixed-precision (bfloat16) support. We trained only bilingual models, a single model for English $\leftrightarrow$ X translation, for a total of 16 models (8 languages, 2 alignment approaches). Each model was trained for 1 epoch. We performed greedy decoding with a maximum of 4,096 generated tokens to maintain consistency across evaluations.

\noindent\textbf{Evaluation Metrics :}  
To evaluate document-level MT performance, we use \textsc{DocCOMET}\footnote{\url{https://github.com/Unbabel/COMET}}, an extension of COMET that incorporates document-level context for improved quality assessment. Specifically, we employ the \texttt{Unbabel/wmt22-comet-da} model from the WMT 2022 Metrics Shared Task, a reference-based COMET model designed for document-level MT evaluation. Alongside COMET, we also report ChrF\footnote{\url{https://github.com/mjpost/sacreBLEU\#chrf--chrf}}, a character-level F-score metric based on n-gram overlaps.

\section{Results and Discussion}
We present results demonstrating the effectiveness of DAC through intrinsic and extrinsic evaluations.

\subsection{Intrinsic Performance of DAC}\label{sec:intrinsic-results}
Table~\ref{tab:intrinsic-eval-results} reports average precision, recall, and F1 scores across granularities and embedding models.

\paragraph{DAC vs. Baselines:}  
DAC consistently achieves the highest precision across all settings, with gains of 15--20\% over pooling-based baselines. While its recall is slightly lower, DAC attains competitive or superior F1 scores, indicating that it produces more precise alignments, a desirable property for high-quality MT training data, where quality is more important than quantity~\cite{ranathunga2024qualitydoesmatterdetailed, zhou2023limaalignment}. Among the baselines, LIDF is the strongest competitor, performing closest to DAC due to its effective combination of length normalization and term weighting. LP performs slightly worse, whereas MP and IDF-based pooling lag considerably behind. \textit{Additionally, in experiments conducted on the CCAligned dataset \cite{el-kishky-etal-2020-ccaligned}, DAC outperforms LIDF across all metrics; see Appendix~\ref{sec:ccaligned_intrinsic-eval} for more details.}

\begin{table*}[!ht]
    \small
    \centering
    \renewcommand{\arraystretch}{1.25}
    \setlength{\tabcolsep}{5pt}

    \definecolor{timeHigh}{RGB}{255,225,225}   
    \definecolor{timeMid}{RGB}{255,245,200}    
    \definecolor{timeLow}{RGB}{195,235,185}    
    \definecolor{timeFast}{RGB}{145,200,130}   

    \definecolor{metric1}{RGB}{236,249,236}    
    \definecolor{metric2}{RGB}{200,234,200}
    \definecolor{metric3}{RGB}{168,220,168}
    \definecolor{metric4}{RGB}{136,205,136}    

    \begin{tabular}{>{\centering\arraybackslash}m{1.5cm}cccccccc}
        \toprule
        \multirow{2}{*}{\textbf{Granularity}} &
        \multicolumn{4}{c}{\textbf{LaBSE} \hspace{3pt} \raisebox{-0.1\height}{\includegraphics[height=10pt]{images/google.png}}} &
        \multicolumn{4}{c}{\textbf{SONAR} \hspace{3pt} \raisebox{-0.2\height}{\includegraphics[height=12pt]{images/meta_logo.png}}} \\
        \cmidrule(lr){2-5} \cmidrule(lr){6-9}
        & \textbf{Time (min)} & \textbf{Precision} & \textbf{Recall} & \textbf{F1} &
          \textbf{Time (min)} & \textbf{Precision} & \textbf{Recall} & \textbf{F1} \\
        \midrule

        G = 1 &
        \cellcolor{timeHigh}47 &
        \cellcolor{metric4}0.9152 &
        \cellcolor{metric4}0.6588 &
        \cellcolor{metric4}0.7635 &
        \cellcolor{timeHigh}115 &
        \cellcolor{metric4}0.9106 &
        \cellcolor{metric4}0.6484 &
        \cellcolor{metric4}0.7550\\

        G = 2 &
        \cellcolor{timeMid}36 &
        \cellcolor{metric3}0.9007 &
        \cellcolor{metric2}0.6411 &
        \cellcolor{metric2}0.7463 &
        \cellcolor{timeMid}96 &
        \cellcolor{metric3}0.8870 &
        \cellcolor{metric2}0.6108 &
        \cellcolor{metric2}0.7210 \\

        G = 4 &
        \cellcolor{timeLow}25 &
        \cellcolor{metric2}0.8936 &
        \cellcolor{metric3}0.6471 &
        \cellcolor{metric2}0.7480 &
        \cellcolor{timeLow}59 &
        \cellcolor{metric2}0.8742 &
        \cellcolor{metric1}0.6005 &
        \cellcolor{metric1}0.7098 \\

        G = 8 &
        \cellcolor{timeFast}18 &
        \cellcolor{metric2}0.8912 &
        \cellcolor{metric2}0.6407 &
        \cellcolor{metric2}0.7428 &
        \cellcolor{timeFast}35 &
        \cellcolor{metric2}0.8732 &
        \cellcolor{metric1}0.6020 &
        \cellcolor{metric2}0.7109 \\
        \bottomrule
    \end{tabular}

    \caption{
Computational time and alignment performance across document granularities (\(G\)) for 150k document pairs on an NVIDIA H100 node. Results for \textsc{LaBSE} and \textsc{SONAR} are reported in wall-clock time (minutes), precision, recall, and F1 score. Color gradients illustrate relative trends: \textcolor{darkred}{\textbf{red}}→\textcolor{darkgreen}{\textbf{green}} indicates slower→faster configurations for time, while performance metrics follow a unified \textcolor{darkgreen}{\textbf{green}} scale. The drop in alignment quality is far smaller than the reduction in computation time, highlighting a strong computational advantage at higher granularities.
}

    \label{efficiency-performance-tradeoff}
\end{table*}

\begin{figure*}[!h] \centering \includegraphics[width=\textwidth]{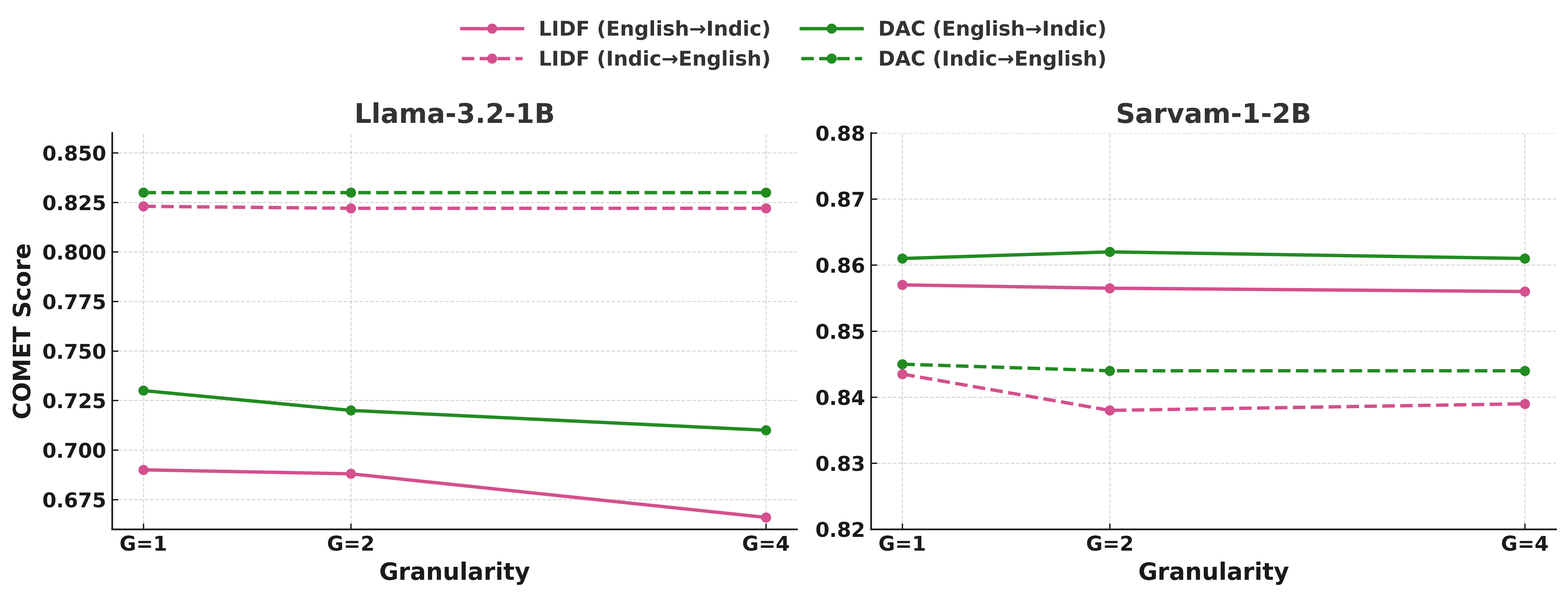} \caption{Extrinsic evaluation of \textsc{DAC} with \textsc{LaBSE} embeddings (\textcolor{darkgreen}{\textbf{green}}) and \textsc{LIDF} with \textsc{SONAR} embeddings (\textcolor{darkpink}{\textbf{pink}}) on \benchname{}. COMET scores averaged across 8 Indic Languages are reported for granularities \(G = 1,2,4\) on English$\rightarrow$Indic (solid lines) and Indic$\rightarrow$English (dashed lines) translation tasks using the \textsc{Llama-3.2-1B} (left) and \textsc{Sarvam-1-2B} (right) models. Appendix~\ref{sec:extrinsic-trends} provides further analysis of extrinsic trends, and detailed per-language COMET scores are presented in Table~\ref{tab:extrinsic_comet_scores_combined}.}
 \label{fig:extrinsic_plot} \end{figure*}

\paragraph{Effect of Granularity.}
As shown in Table~\ref{efficiency-performance-tradeoff}, alignment performance is highest at the finest granularity (\(G = 1\)) for both LaBSE and SONAR, with only a modest decline as chunk size increases. Beyond \(G = 2\), variations across performance metrics remain minimal, indicating that coarser segmentations preserve much of the alignment quality. In contrast, computational time decreases sharply from 47 to 18 minutes for LaBSE and from 115 to 35 minutes for SONAR as \(G\) increases from 1 to 8, resulting in roughly a 2–3× reduction in runtime. This efficiency gain stems from a combinatorial reduction in the candidate search space: at higher granularities, each document is represented by fewer, longer chunks, leading to substantially fewer embedding computations and pairwise similarity operations during retrieval.

This trade-off underscores an important consideration for large-scale document alignment. While fine-grained (\(G = 1\)) configurations yield slightly better alignment quality, their computational cost becomes impractical at scale. Coarser granularities thus offer a balanced solution, maintaining competitive alignment performance while enabling significantly faster processing, offering a favorable efficiency--performance trade-off .

\paragraph{Effect of Embedding Model.}
Across embedding models, \textsc{SONAR} performs better with pooling-based methods, whereas \textsc{LaBSE} achieves higher performance with DAC. This contrast reflects their underlying training objectives: \textsc{SONAR} is trained with mean-pooled sentence representations, making it naturally compatible with aggregation-based alignment, while \textsc{LaBSE} relies on the [CLS]-token embedding, which encodes context from the entire sequence while remaining sensitive to token-level structure. Consequently, \textsc{LaBSE} benefits from methods that preserve token-level information, whereas \textsc{SONAR} aligns more effectively with pooling strategies that emphasize global semantics. See Appendix~\ref{sec:intrinsic-trends} for a detailed analysis of intrinsic performance trends.

\subsection{Extrinsic Performance of DAC}\label{sec:extrinsic-results}

Based on the intrinsic evaluations, the optimal configurations for document alignment on our \textsc{Pralekha} dataset are \textsc{LaBSE} with DAC at granularities \(G = 1\), \(G = 2\), and \(G = 4\), followed closely by \textsc{SONAR} with LIDF at the same granularities. These configurations were chosen solely on the basis of alignment performance metrics, without accounting for computational cost. In this section, we evaluate their extrinsic performance using the experimental setup described in Section~\ref{sec:extrinsic-eval}, to examine whether the intrinsic alignment trends translate into measurable downstream gains.

Table~\ref{tab:extrinsic_comet_scores_combined} reports COMET scores for extrinsic evaluation, comparing DAC with LIDF (the strongest baseline). \emph{DAC consistently outperforms LIDF across all granularities, translation directions, and fine-tuned models}, demonstrating that higher-precision alignments yield cleaner parallel data and result in measurable downstream improvements in translation quality. Detailed per-language COMET scores are presented in Appendix~\ref{sec:extrinsic-trends}.

Differences in MT performance across granularities are relatively small and align closely with intrinsic performance trends. The strong results for \(G = 2\) and \(G = 4\) suggest that moderate chunking can substantially reduce computational cost while maintaining translation quality. This suggests that coarse-grained segmentation balances alignment efficiency and performance in large-scale settings.

Across models, Sarvam-based systems outperform their LLaMA-based counterparts, particularly for English$\to$Indic translation. This is likely due to Sarvam’s tokenizer being optimized for Indic languages, reducing token fertility and generation burden. Conversely, Indic$\to$English translations achieve consistently higher scores, reflecting the greater morphological and syntactic complexity of Indic languages, which makes it more challenging.

 Overall, performance differences remain modest but consistent across settings. Extrinsic results confirm that DAC-aligned corpora at \(G\) = 2 \& 4 achieve strong translation quality. However, it is worth noting that intrinsic performance does not always translate to downstream performance. Appendix~\ref{sec:extrinsic-trends} provides further discussion of these trends, and Table~\ref{tab:extrinsic_chrf_scores_combined} reports corresponding ChrF scores, which follow patterns similar to the COMET results.

\section{Conclusion}
\label{sec:conclusion}

Mining parallel document pairs for document-level MT is challenging because existing CLDA methods often rely on metadata or pooled sentence embeddings due to the limited context windows of embedding models. Moreover, sentence-level alignment introduces a combinatorially large search space, leading to high computational costs that become impractical at scale. To address these challenges for Indic languages, we introduced \benchname{}, a large-scale benchmark comprising over 3 million aligned document pairs across 11 Indic languages and English. Furthermore, we proposed the Document Alignment Coefficient (DAC), a novel metric for fine-grained document alignment. Unlike pooling-based methods, DAC aligns documents by matching smaller chunks and computes similarity as the ratio of aligned chunks to the average number of chunks in a pair. Intrinsic evaluation shows that our chunk-based approach is 2–3× faster while maintaining competitive performance, and that DAC achieves substantial gains over pooling-based baselines. Extrinsic evaluation further demonstrates that document-level MT models trained on DAC-aligned pairs consistently outperform those trained using baseline alignment methods. These results highlight DAC’s effectiveness for parallel document mining, balancing performance with computational efficiency.
By releasing the \benchname{} dataset and CLDA evaluation framework, we aim to facilitate future research on scalable and reliable document-level alignment and to advance the development of high-quality document-level MT systems for Indic languages.

\section*{Limitations}
\label{sec:limitations}

While \benchname{} and the proposed chunk-based alignment approach advance research in cross-lingual document alignment, some limitations remain. Our dataset covers 11 Indic languages and English, so the findings may not generalize to other language families with different linguistic properties. The documents in \benchname{} are primarily drawn from structured, high-quality sources, making the benchmark less representative of noisy web data that includes informal language, OCR errors, or code-switching. In addition, the benchmark mostly contains documents of moderate length, leaving performance on very short or long documents, common in other domains, largely untested.
Because the dataset is domain-specific, MT models trained solely on it may overfit and exhibit reduced generalization, or even hallucinate when evaluated on out-of-domain data. DAC may also be less robust when translations are semantically equivalent but highly paraphrased. Finally, our extrinsic evaluation does not fully establish how improvements in alignment quality affect document-level MT performance, as current MT metrics capture discourse-level phenomena only to a limited extent.
Future work should expand CLDA evaluation to more diverse language families, noisier domains, and a wider range of document lengths. It should also explore alignment strategies that are more robust to paraphrasing and develop document-level MT metrics that better correlate with alignment quality.

\section*{Ethics}
\label{sec:ethics}

All datasets used in this paper are available under permissible licenses, and we adhere strictly to their intended usage, maintaining full compliance with licensing requirements. Additionally, the code used for our evaluation framework will be made publicly available under the MIT License.\footnote{\url{https://opensource.org/licenses/MIT}} Generative AI systems were used solely to assist with the language of this paper, specifically for paraphrasing, spell-checking, polishing the authors' original text, and generating boilerplate code.

\section*{Acknowledgements}
\label{sec:acknowledgements}
We would like to thank EkStep Foundation and
Nilekani Philanthropies for their generous grant
towards building datasets, models, tools, and other
resources for Indian languages. This work was partly supported by JSPS KAKENHI Grant-in-Aid for Early-Career Scientists 25K21290.

\bibliography{custom}

@article{resnik-smith-2003-web,
    title = {{The Web as a Parallel Corpus}},
    author = {Resnik, Philip and Smith, Noah A.},
    journal = {American Journal of Computational Linguistics},
    volume = {29},
    number = {3},
    year = {2003},
    url = {https://aclanthology.org/J03-3002/},
    doi = {10.1162/089120103322711578},
    pages = {349--380}
}

@misc{gala2023indictrans2highqualityaccessiblemachine,
      title={{IndicTrans2: Towards High-Quality and Accessible Machine Translation Models for all 22 Scheduled Indian Languages}}, 
      author={Jay Gala and Pranjal A. Chitale and Raghavan AK and Varun Gumma and Sumanth Doddapaneni and Aswanth Kumar and Janki Nawale and Anupama Sujatha and Ratish Puduppully and Vivek Raghavan and Pratyush Kumar and Mitesh M. Khapra and Raj Dabre and Anoop Kunchukuttan},
      year={2023},
      eprint={2305.16307},
      archivePrefix={arXiv},
      primaryClass={cs.CL},
      url={https://arxiv.org/abs/2305.16307}, 
}

@misc{nllbteam2022languageleftbehindscaling,
      title={{No Language Left Behind: Scaling Human-Centered Machine Translation}}, 
      author={NLLB Team and Marta R. Costa-jussà and James Cross and Onur Çelebi and Maha Elbayad and Kenneth Heafield and Kevin Heffernan and Elahe Kalbassi and Janice Lam and Daniel Licht and Jean Maillard and Anna Sun and Skyler Wang and Guillaume Wenzek and Al Youngblood and Bapi Akula and Loic Barrault and Gabriel Mejia Gonzalez and Prangthip Hansanti and John Hoffman and Semarley Jarrett and Kaushik Ram Sadagopan and Dirk Rowe and Shannon Spruit and Chau Tran and Pierre Andrews and Necip Fazil Ayan and Shruti Bhosale and Sergey Edunov and Angela Fan and Cynthia Gao and Vedanuj Goswami and Francisco Guzmán and Philipp Koehn and Alexandre Mourachko and Christophe Ropers and Safiyyah Saleem and Holger Schwenk and Jeff Wang},
      year={2022},
      eprint={2207.04672},
      archivePrefix={arXiv},
      primaryClass={cs.CL},
      url={https://arxiv.org/abs/2207.04672}, 
}

@misc{xie2024chunkalignselectsimple,
      title={{Chunk, Align, Select: A Simple Long-sequence Processing Method for Transformers}}, 
      author={Jiawen Xie and Pengyu Cheng and Xiao Liang and Yong Dai and Nan Du},
      year={2024},
      eprint={2308.13191},
      archivePrefix={arXiv},
      primaryClass={cs.CL},
      url={https://arxiv.org/abs/2308.13191}, 
}

@inproceedings{sannigrahi-etal-2023-best,
    title = "Are the Best Multilingual Document Embeddings simply Based on Sentence Embeddings?",
    author = "Sannigrahi, Sonal  and
      van Genabith, Josef  and
      Espa{\~n}a-Bonet, Cristina",
    editor = "Vlachos, Andreas  and
      Augenstein, Isabelle",
    booktitle = "Findings of the Association for Computational Linguistics: EACL 2023",
    month = may,
    year = "2023",
    address = "Dubrovnik, Croatia",
    publisher = "Association for Computational Linguistics",
    url = "https://aclanthology.org/2023.findings-eacl.174/",
    doi = "10.18653/v1/2023.findings-eacl.174",
    pages = "2306--2316",
}

@inproceedings{schwenk-douze-2017-learning,
  title={{Learning Joint Multilingual Sentence Representations with Neural Machine Translation}},
  author={Schwenk, Holger and Douze, Matthijs},
  booktitle={Proceedings of the 2nd Workshop on Representation Learning for {NLP}},
  year={2017},
  address={Vancouver, Canada},
  publisher={Association for Computational Linguistics},
  pages={157--167},
  doi={10.18653/v1/W17-2619},
  url={https://aclanthology.org/W17-2619/}
}

@inproceedings{pal-etal-2024-document,
    title = {{Document-Level Machine Translation with Large-Scale Public Parallel Corpora}},
    author = "Pal, Proyag  and
      Birch, Alexandra  and
      Heafield, Kenneth",
    editor = "Ku, Lun-Wei  and
      Martins, Andre  and
      Srikumar, Vivek",
    booktitle = "Proceedings of the 62nd Annual Meeting of the Association for Computational Linguistics (Volume 1: Long Papers)",
    month = aug,
    year = "2024",
    address = "Bangkok, Thailand",
    publisher = "Association for Computational Linguistics",
    url = "https://aclanthology.org/2024.acl-long.712/",
    doi = "10.18653/v1/2024.acl-long.712",
    pages = "13185--13197",
}

@article{ramesh-etal-2022-samanantar,
  title={{Samanantar: The Largest Publicly Available Parallel Corpora Collection for 11 {I}ndic Languages}},
  author={Ramesh, Gowtham and Doddapaneni, Sumanth and Bheemaraj, Aravinth and Jobanputra, Mayank and AK, Raghavan and Sharma, Ajitesh and Sahoo, Sujit and Diddee, Harshita and J, Mahalakshmi and Kakwani, Divyanshu and Kumar, Navneet and Pradeep, Aswin and Nagaraj, Srihari and Deepak, Kumar and Raghavan, Vivek and Kunchukuttan, Anoop and Kumar, Pratyush and Khapra, Mitesh Shantadevi},
  journal={Transactions of the Association for Computational Linguistics},
  volume={10},
  pages={145--162},
  year={2022},
  publisher={MIT Press},
  doi={10.1162/tacl_a_00452},
  url={https://aclanthology.org/2022.tacl-1.9/}
}

@inproceedings{khan-etal-2024-indicllmsuite,
  title     = {{IndicLLMSuite: A Blueprint for Creating Pre-training and Fine-Tuning Datasets for Indian Languages}},
  author    = {Khan, Mohammed Safi Ur Rahman and Mehta, Priyam and Sankar, Ananth and Kumaravelan, Umashankar and Doddapaneni, Sumanth and Suriyaprasaad, G and Balan, Varun G and Jain, Sparsh and Kunchukuttan, Anoop and Kumar, Pratyush and Dabre, Raj and Khapra, Mitesh M.},
  booktitle = {Proceedings of the 62nd Annual Meeting of the Association for Computational Linguistics (Volume 1: Long Papers)},
  year      = {2024},
  pages     = {15831--15879},
  publisher = {Association for Computational Linguistics},
  url       = {https://aclanthology.org/2024.acl-long.843},
  doi       = {10.18653/v1/2024.acl-long.843}
}

@misc{duquenne2023sonarsentencelevelmultimodallanguageagnostic,
      title={{SONAR: Sentence-Level Multimodal and Language-Agnostic Representations}}, 
      author={Paul-Ambroise Duquenne and Holger Schwenk and Benoît Sagot},
      year={2023},
      eprint={2308.11466},
      archivePrefix={arXiv},
      primaryClass={cs.CL},
      url={https://arxiv.org/abs/2308.11466}, 
}

@inproceedings{banon-etal-2020-paracrawl,
    title = {{{P}ara{C}rawl: Web-Scale Acquisition of Parallel Corpora}},
    author = "Ba{\~n}{\'o}n, Marta  and
      Chen, Pinzhen  and
      Haddow, Barry  and
      Heafield, Kenneth  and
      Hoang, Hieu  and
      Espl{\`a}-Gomis, Miquel  and
      Forcada, Mikel L.  and
      Kamran, Amir  and
      Kirefu, Faheem  and
      Koehn, Philipp  and
      Ortiz Rojas, Sergio  and
      Pla Sempere, Leopoldo  and
      Ram{\'i}rez-S{\'a}nchez, Gema  and
      Sarr{\'i}as, Elsa  and
      Strelec, Marek  and
      Thompson, Brian  and
      Waites, William  and
      Wiggins, Dion  and
      Zaragoza, Jaume",
    editor = "Jurafsky, Dan  and
      Chai, Joyce  and
      Schluter, Natalie  and
      Tetreault, Joel",
    booktitle = "Proceedings of the 58th Annual Meeting of the Association for Computational Linguistics",
    month = jul,
    year = "2020",
    address = "Online",
    publisher = "Association for Computational Linguistics",
    url = "https://aclanthology.org/2020.acl-main.417/",
    doi = "10.18653/v1/2020.acl-main.417",
    pages = "4555--4567"
}

@article{el-kishky-guzman-2020-massively,
  title={{Massively Multilingual Document Alignment with Cross-lingual Sentence-Mover's Distance}},
  author={El-Kishky, Ahmed and Guzm{\'a}n, Francisco},
  journal={arXiv preprint arXiv:2002.00761},
  year={2020},
  url={https://arxiv.org/abs/2002.00761}
}

@misc{schwenk2020ccmatrixminingbillionshighquality,
      title={{CCMatrix: Mining Billions of High-Quality Parallel Sentences on the WEB}}, 
      author={Holger Schwenk and Guillaume Wenzek and Sergey Edunov and Edouard Grave and Armand Joulin},
      year={2020},
      eprint={1911.04944},
      archivePrefix={arXiv},
      primaryClass={cs.CL},
      url={https://arxiv.org/abs/1911.04944}, 
}

@misc{ranathunga2024qualitydoesmatterdetailed,
      title={{Quality Does Matter: A Detailed Look at the Quality and Utility of Web-Mined Parallel Corpora}}, 
      author={Surangika Ranathunga and Nisansa de Silva and Menan Velayuthan and Aloka Fernando and Charitha Rathnayake},
      year={2024},
      eprint={2402.07446},
      archivePrefix={arXiv},
      primaryClass={cs.CL},
      url={https://arxiv.org/abs/2402.07446}, 
}

@inproceedings{feng-etal-2022-language,
  title={{Language-Agnostic BERT Sentence Embedding}},
  author={Feng, Fangxiaoyu and Yang, Yinfei and Cer, Daniel and Arivazhagan, Naveen and Wang, Wei},
  booktitle={Proceedings of the 60th Annual Meeting of the Association for Computational Linguistics (ACL)},
  year={2022},
  pages={878--894},
  publisher={Association for Computational Linguistics},
  url={https://aclanthology.org/2022.acl-long.62},
  doi={10.18653/v1/2022.acl-long.62},
}

@inproceedings{artetxe-schwenk-2019-margin,
  title={{Margin-based Parallel Corpus Mining with Multilingual Sentence Embeddings}},
  author={Artetxe, Mikel and Schwenk, Holger},
  booktitle={Proceedings of the 57th Annual Meeting of the Association for Computational Linguistics},
  pages={3197--3203},
  year={2019},
  month={July},
  address={Florence, Italy},
  publisher={Association for Computational Linguistics},
  url={https://aclanthology.org/P19-1309},
  doi={10.18653/v1/P19-1309}
}

@inproceedings{schwenk-etal-2021-wikimatrix,
  title={{WikiMatrix: Mining 135 Million Parallel Sentences from Wikipedia}},
  author={Schwenk, Holger and Chaudhary, Vishrav and Sun, Shuo and Gong, Hongyu and Guzm{\'a}n, Francisco},
  booktitle={Proceedings of the 16th Conference of the European Chapter of the Association for Computational Linguistics},
  year={2021},
  pages={1625--1639},
  publisher={Association for Computational Linguistics},
  url={https://aclanthology.org/2021.eacl-main.115},
  doi={10.18653/v1/2021.eacl-main.115},
}

@inproceedings{koehn-2005-europarl,
  title={{Europarl: A Parallel Corpus for Statistical Machine Translation}},
  author={Koehn, Philipp},
  booktitle={Proceedings of Machine Translation Summit X: Papers},
  year={2005},
  address={Phuket, Thailand},
  pages={79--86},
  url={https://aclanthology.org/2005.mtsummit-papers.11},
}

@inproceedings{abdul-rauf-schwenk-2009-improving,
  title={{On the Use of Comparable Corpora to Improve SMT Performance}},
  author={Abdul-Rauf, Sadaf and Schwenk, Holger},
  booktitle={Proceedings of the 12th Conference of the European Chapter of the ACL (EACL 2009)},
  year={2009},
  pages={16--23},
  publisher={Association for Computational Linguistics},
  url={https://aclanthology.org/E09-1003},
}

@inproceedings{do-etal-2009-mining,
  title={{Mining a Comparable Text Corpus for a Vietnamese-French Statistical Machine Translation System}},
  author={Do, Thi-Ngoc-Diep and Le, Viet-Bac and Bigi, Brigitte and Besacier, Laurent and Castelli, Eric},
  booktitle={Proceedings of the Fourth Workshop on Statistical Machine Translation},
  year={2009},
  pages={165--172},
  publisher={Association for Computational Linguistics},
  url={https://aclanthology.org/W09-0430},
}

@inproceedings{siripragada-etal-2020-multilingual,
  title={{A Multilingual Parallel Corpora Collection Effort for Indian Languages}},
  author={Siripragada, Shashank and Philip, Jerin and Namboodiri, Vinay P. and Jawahar, C V},
  booktitle={Proceedings of the Twelfth Language Resources and Evaluation Conference},
  year={2020},
  pages={3743--3751},
  publisher={European Language Resources Association},
  url={https://aclanthology.org/2020.lrec-1.462},
}

@misc{zhu2024longembedextendingembeddingmodels,
      title={{LongEmbed: Extending Embedding Models for Long Context Retrieval}}, 
      author={Dawei Zhu and Liang Wang and Nan Yang and Yifan Song and Wenhao Wu and Furu Wei and Sujian Li},
      year={2024},
      eprint={2404.12096},
      archivePrefix={arXiv},
      primaryClass={cs.CL},
      url={https://arxiv.org/abs/2404.12096}, 
}

@inproceedings{el-kishky-etal-2020-ccaligned,
  title={{CCAligned: A Massive Collection of Cross-Lingual Web-Document Pairs}},
  author={El-Kishky, Ahmed and Chaudhary, Vishrav and Guzm{\'a}n, Francisco and Koehn, Philipp},
  booktitle={Proceedings of the 2020 Conference on Empirical Methods in Natural Language Processing},
  year={2020},
  pages={5960--5969},
  publisher={Association for Computational Linguistics},
  url={https://aclanthology.org/2020.emnlp-main.480},
  doi={10.18653/v1/2020.emnlp-main.480},
}

@inproceedings{guo-etal-2019-hierarchical,
  title={{Hierarchical Document Encoder for Parallel Corpus Mining}},
  author={Guo, Mandy and Yang, Yinfei and Stevens, Keith and Cer, Daniel and Ge, Heming and Sung, Yun-Hsuan and Strope, Brian and Kurzweil, Ray},
  booktitle={Proceedings of the Fourth Conference on Machine Translation (Volume 1: Research Papers)},
  year={2019},
  pages={64--72},
  publisher={Association for Computational Linguistics},
  url={https://aclanthology.org/W19-5207},
  doi={10.18653/v1/W19-5207},
}

@inproceedings{kakwani-etal-2020-indicnlpsuite,
  title={{IndicNLPSuite: Monolingual Corpora, Evaluation Benchmarks and Pre-trained Multilingual Language Models for Indian Languages}},
  author={Kakwani, Divyanshu and Kunchukuttan, Anoop and Golla, Satish and N.C., Gokul and Bhattacharyya, Avik and Khapra, Mitesh M. and Kumar, Pratyush},
  booktitle={Findings of the Association for Computational Linguistics: EMNLP 2020},
  year={2020},
  pages={4948--4961},
  publisher={Association for Computational Linguistics},
  url={https://aclanthology.org/2020.findings-emnlp.445},
  doi={10.18653/v1/2020.findings-emnlp.445},
}

@inproceedings{madhani-etal-2023-bhasa-abhijnaanam,
    title = {{{B}hasa-{A}bhijnaanam: {N}ative-script and romanized {L}anguage {I}dentification for 22 {I}ndic {l}anguages}},
    author = "Madhani, Yash and Khapra, Mitesh M. and Kunchukuttan, Anoop",
    booktitle = "Proceedings of the 61st Annual Meeting of the Association for Computational Linguistics (Volume 2: Short Papers)",
    month = jul,
    year = "2023",
    address = "Toronto, Canada",
    editor = "Anna Rogers and Jordan Boyd-Graber and Naoaki Okazaki",
    publisher = "Association for Computational Linguistics",
    url = "https://aclanthology.org/2023.acl-short.71",
    doi = "10.18653/v1/2023.acl-short.71",
    pages = "816--826",
}

@misc{zhou2023limaalignment,
      title={{LIMA: Less Is More for Alignment}}, 
      author={Chunting Zhou and Pengfei Liu and Puxin Xu and Srini Iyer and Jiao Sun and Yuning Mao and Xuezhe Ma and Avia Efrat and Ping Yu and Lili Yu and Susan Zhang and Gargi Ghosh and Mike Lewis and Luke Zettlemoyer and Omer Levy},
      year={2023},
      eprint={2305.11206},
      archivePrefix={arXiv},
      primaryClass={cs.CL},
      url={https://arxiv.org/abs/2305.11206}, 
}

\newpage \appendix
\newpage \newpage {\noindent\LARGE{\textbf{Appendix}}}
\label{sec:appendix}

\section{Intrinsic Performance Trends in CLDA}
\label{sec:intrinsic-trends}

This section analyzes the Precision, Recall, and F1 score trends of various Cross-Lingual Document Alignment (CLDA) methods based on their intrinsic performance on \benchname{}, as shown in Table~\ref{tab:intrinsic-eval-results}.

\subsection{Impact of Granularities}

\begin{figure}[!h] \centering \includegraphics[width=\linewidth]{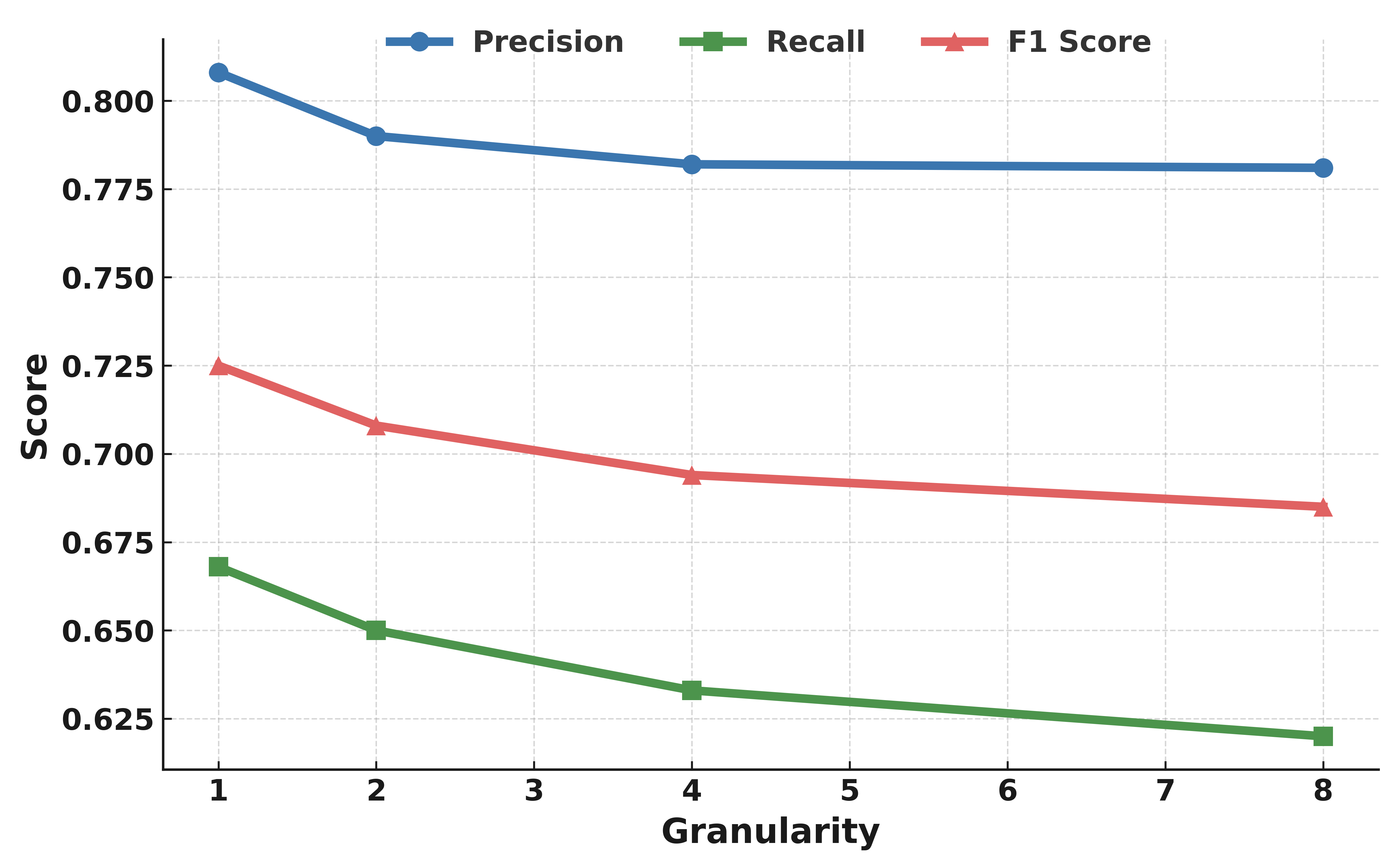} \caption{Impact of Granularity on the Intrinsic Performance of CLDA techniques on \benchname{}.} \label{fig:intrinsic-granularity-trend} \end{figure}

As shown in Figure~\ref{fig:intrinsic-granularity-trend}, increasing granularity is associated with a gradual reduction in precision, recall, and F1 score. The decline is most pronounced for recall, suggesting that fewer relevant matches are retrieved as segmentation becomes finer. Precision, however, remains relatively stable across granularities. These trends indicate that while finer segmentation can provide more detailed units for alignment, it may also reduce the amount of contextual information available to embedding models, which are typically optimized for shorter text spans.

\subsection{Impact of Embedding Models}

\begin{figure}[!h] \centering \includegraphics[width=\linewidth]{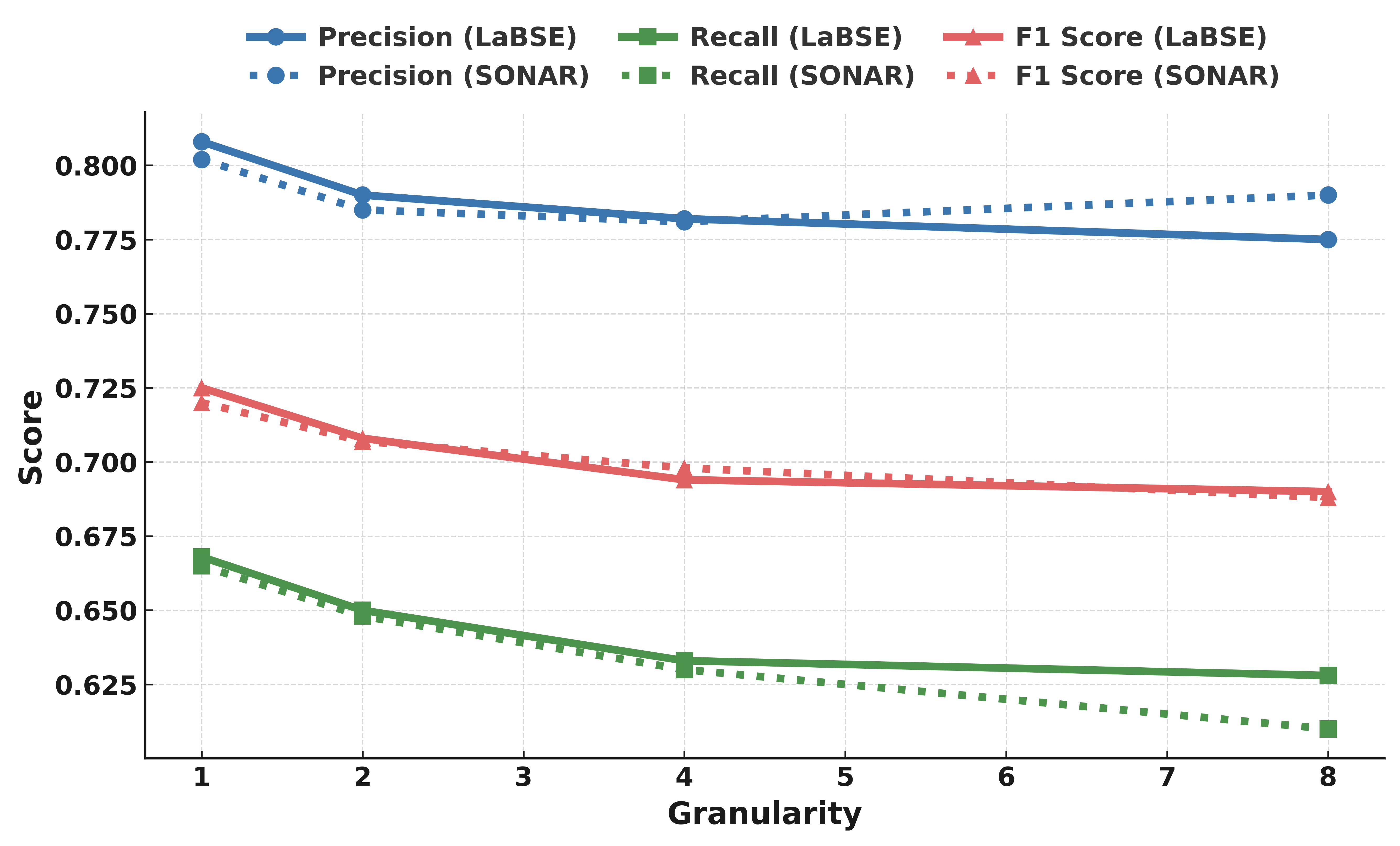} \caption{Impact of Embedding Models on the Intrinsic Performance of CLDA techniques on \benchname{}.} \label{fig:intrinsic-embedding-trend} \end{figure}

Figure~\ref{fig:intrinsic-embedding-trend} shows how increasing granularity affects the performance of both LaBSE and SONAR across precision, recall, and F1 score. SONAR tends to maintain slightly higher precision, indicating more accurate alignments, whereas LaBSE achieves higher recall, retrieving a larger number of matches. At finer granularities, both models show a gradual decrease in overall performance, with SONAR being relatively more stable in precision. LaBSE achieves a higher F1 score overall, reflecting a more balanced trade-off between precision and recall. These results suggest that SONAR may be better suited for scenarios where precision is prioritized, while LaBSE can be advantageous for broader retrieval needs.

\subsection{Impact of DAC Threshold}

\begin{figure}[!h]
    \centering
    \includegraphics[width=\linewidth]{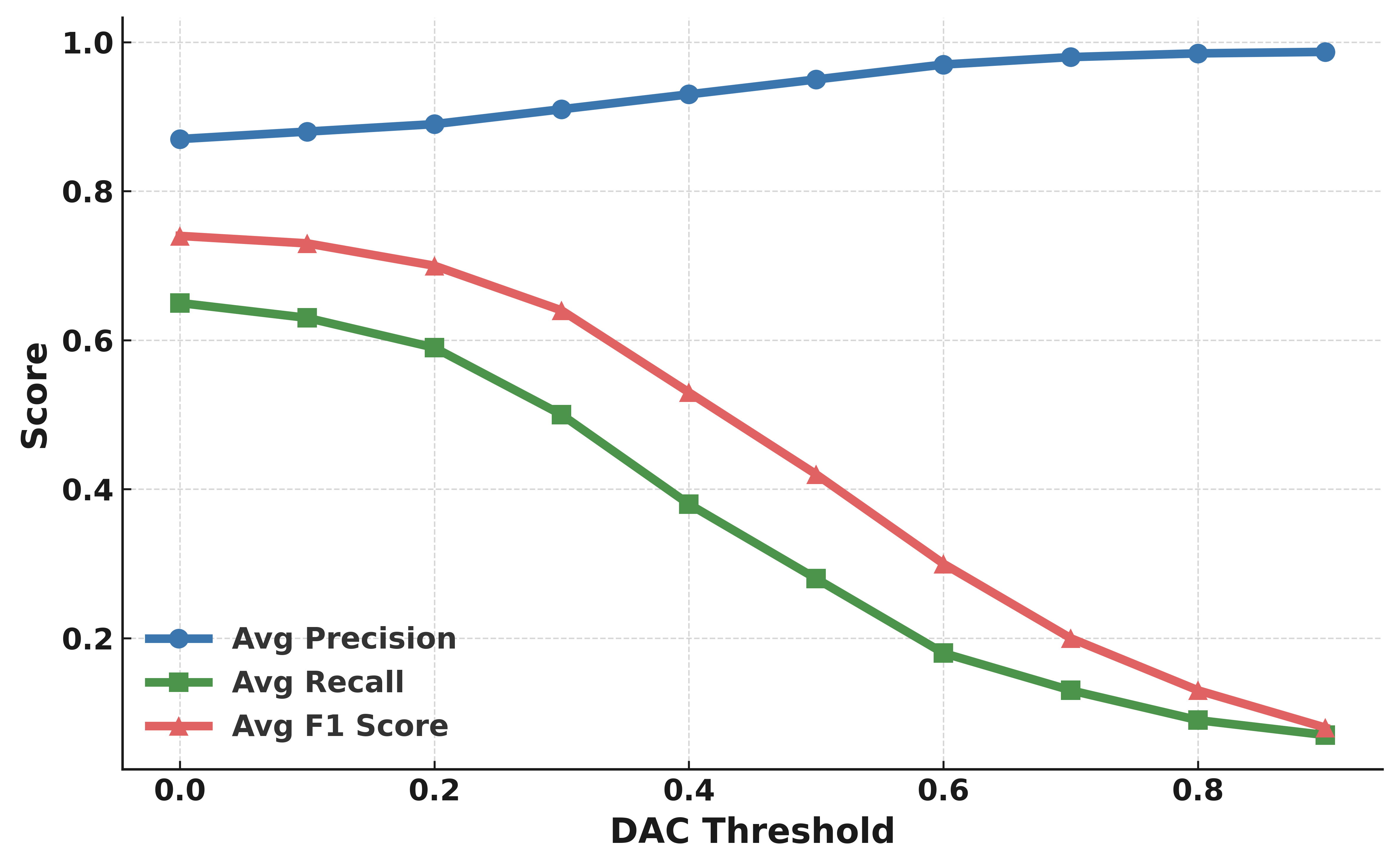}
    \caption{Impact of DAC Threshold on the Intrinsic Performance of CLDA techniques on \benchname{}.}
    \label{fig:dac_curve}
\end{figure}

Figure~\ref{fig:dac_curve} illustrates the trends of precision, recall, and F1 score across different DAC thresholds, highlighting the trade-offs between precision and recall. Increasing the DAC threshold leads to higher precision but lower recall and F1 scores in all settings. We adopt a threshold of 0.1 for both intrinsic and extrinsic evaluations, as it provides a better balance between quality and yield. Table~\ref{tab:dac-intrinsic-eval-results} compares the intrinsic performance of DAC at thresholds 0.1 and 0.5.

\begin{table*}[!ht]
\footnotesize
\centering
\begin{tabular}{llcccccccc}
\toprule
\multirow{2}{*}{\textbf{Metrics}} & \multirow{2}{*}{\textbf{Methods}} 
  & \multicolumn{4}{c}{\textbf{LaBSE} \hspace{3pt} \raisebox{-0.1\height}{\includegraphics[height=10pt]{images/google.png}}}
  & \multicolumn{4}{c}{\textbf{SONAR} \hspace{3pt} \raisebox{-0.2\height}{\includegraphics[height=12pt]{images/meta_logo.png}}} \\
\cmidrule(lr){3-6} \cmidrule(lr){7-10}
  &   & \textbf{G = 1} & \textbf{G = 2} & \textbf{G = 4} & \textbf{G = 8} 
      & \textbf{G = 1} & \textbf{G = 2} & \textbf{G = 4} & \textbf{G = 8} \\
\midrule
\multirow{2}{*}{Precision} 
  & DAC (0.1)  & 0.9152 & 0.9007 & 0.8936 & 0.8912 & 0.9106 & 0.8870 & 0.8742 & 0.8732 \\
  & DAC (0.5)  & \textbf{0.9555} & \textbf{0.9617} & \textbf{0.9570} & \textbf{0.9404} & \textbf{0.9546} & \textbf{0.9600} & \textbf{0.9480} & \textbf{0.9307} \\
\midrule
\multirow{2}{*}{Recall} 
  & DAC (0.1)  & \textbf{0.6588} & \textbf{0.6411} & \textbf{0.6471} & \textbf{0.6407} & \textbf{0.6484} & \textbf{0.6108} & \textbf{0.6005} & \textbf{0.6020} \\
  & DAC (0.5)  & 0.2930 & 0.2568 & 0.3104 & 0.3960 & 0.2530 & 0.2005 & 0.2542 & 0.3297 \\
\midrule
\multirow{2}{*}{F1 Score} 
  & DAC (0.1)  & \textbf{0.7635} & \textbf{0.7463} & \textbf{0.7480} & \textbf{0.7428} & \textbf{0.7550} & \textbf{0.7210} & \textbf{0.7098} & \textbf{0.7109} \\
  & DAC (0.5)  & 0.4368 & 0.3915 & 0.4564 & 0.5486 & 0.3896 & 0.3229 & 0.3931 & 0.4814 \\
\bottomrule
\end{tabular}
\caption{Comparison of DAC threshold performance (0.1 and 0.5) in the intrinsic evaluation on \benchname{}. \textbf{Bold} values indicate the best score for each granularity.}
\label{tab:dac-intrinsic-eval-results}
\end{table*}

\begin{figure*}[!h] \centering \includegraphics[width=\textwidth]{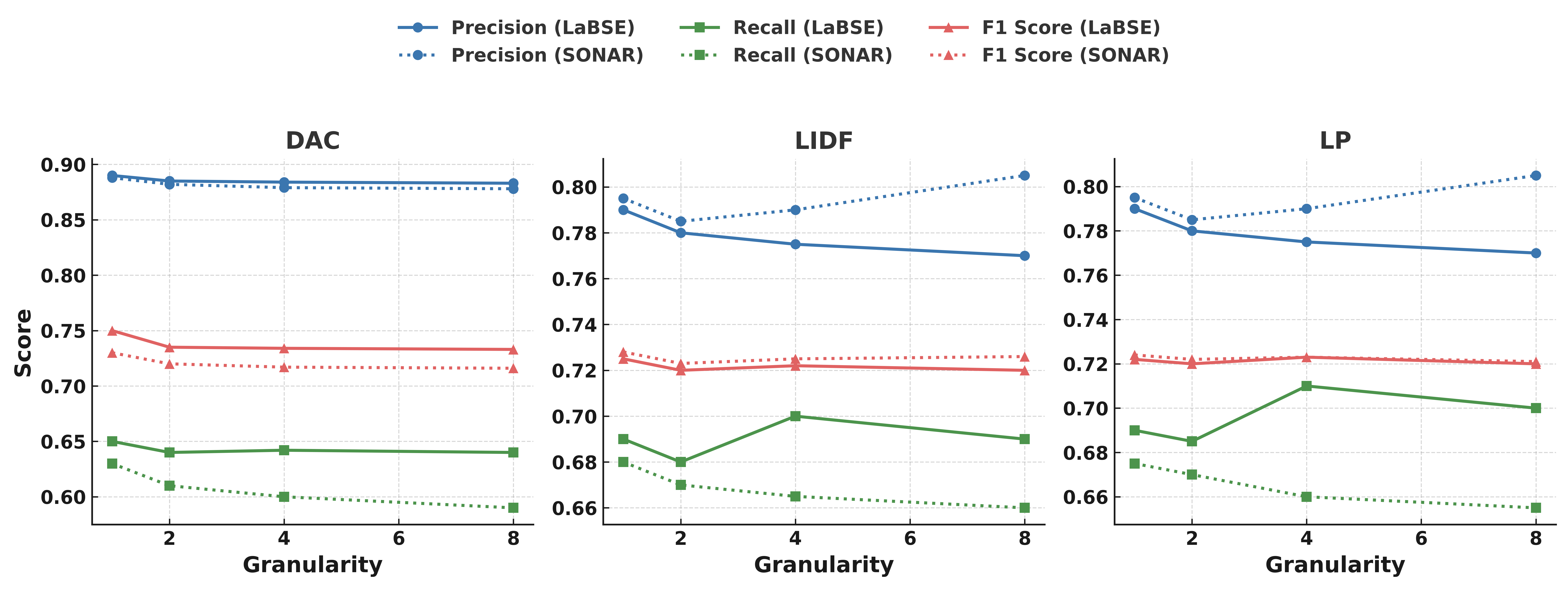} \caption{Intrinsic evaluation performance of DAC, LP, and LIDF on \benchname{}.} \label{fig:intrinsic-methods-trend} \end{figure*}

\subsection{DAC vs. Baseline Methods}
We compare our DAC approach with the two strongest pooling-based baselines, Length Pooling (LP) and Length-IDF (LIDF), as shown in Table~\ref{tab:intrinsic-eval-results}.
Figure~\ref{fig:intrinsic-methods-trend} highlights that DAC consistently delivers the highest precision across all granularities for both LaBSE and SONAR. Although LP and LIDF yield slightly higher recall, DAC achieves a better balance between precision and recall, reflected in its F1 scores. Moreover, DAC shows minimal degradation in precision as granularity increases, demonstrating its robustness compared to the baselines. These trends confirm that DAC more effectively leverages contextual information, making it the strongest overall alignment method.

\begin{figure*}[!h] \centering \includegraphics[width=\textwidth]{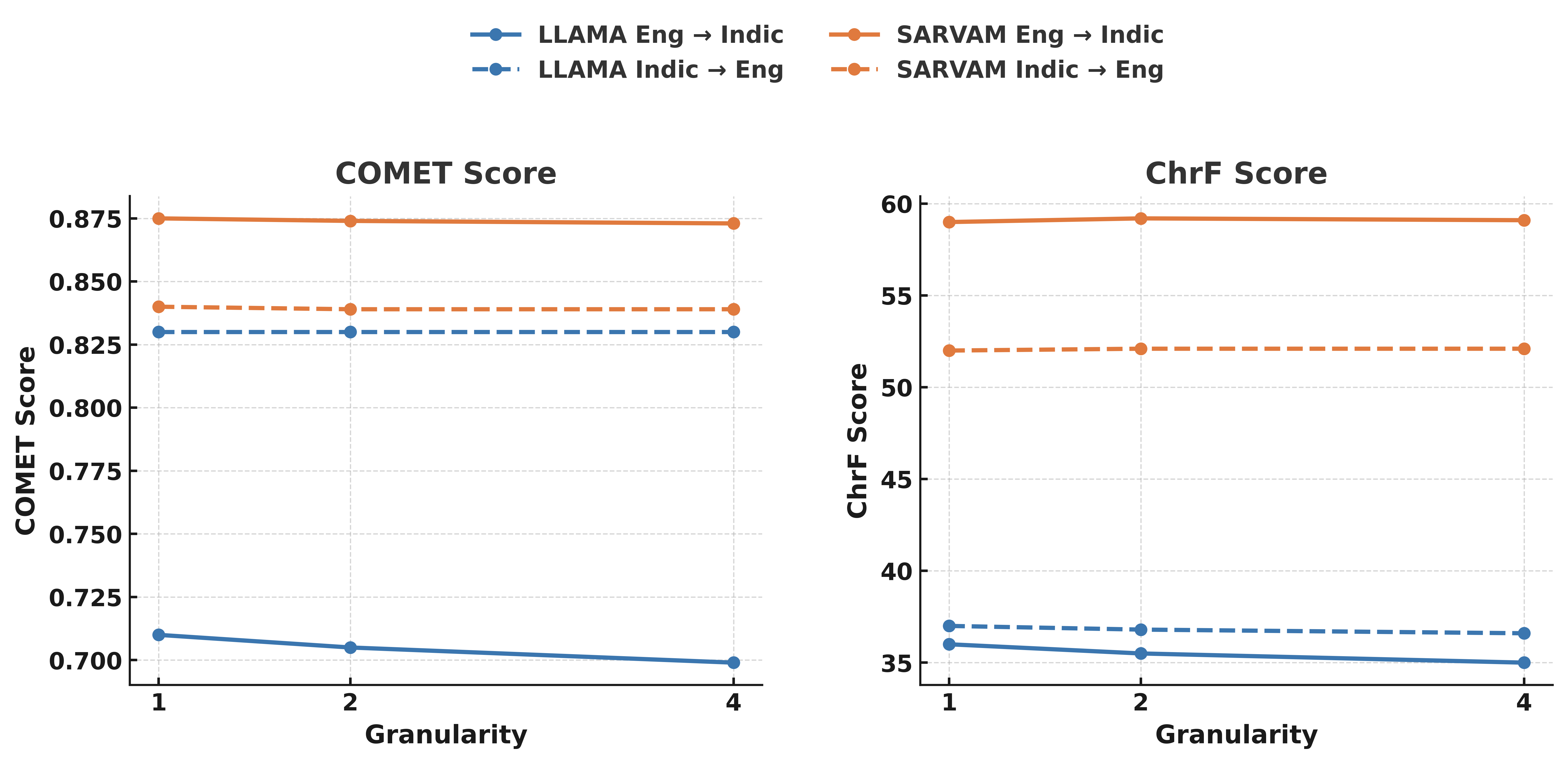} \caption{Extrinsic Evaluation of \textsc{Llama-3.2-1B} (\textcolor{darkpastelblue}{\textbf{blue}}) and \textsc{Sarvam-1-2B} (\textcolor{darkpastelorange}{\textbf{orange}}) on \benchname{}, reporting COMET (left) and ChrF (right) scores averaged across 8 Indic Languages on \benchname{} across granularities \(G = 1,2,4\) on English$\rightarrow$Indic (solid lines) and Indic$\rightarrow$English (dashed lines) translation tasks.}
 \label{fig:extrinsic_comet-chrf_plot} \end{figure*}

\section{Intrinsic Evaluation on \textsc{CCAligned}}
\label{sec:ccaligned_intrinsic-eval}

\begingroup
\setlength{\tabcolsep}{5pt}

\begin{table*}[!ht]
\scriptsize
\centering
\resizebox{\textwidth}{!}{%
\begin{tabular}{llccc ccc ccc}
\toprule
\multirowcell{2}{\textbf{Language}} & \multirowcell{2}{\textbf{Method}} 
& \multicolumn{3}{c}{\textbf{Precision}} 
& \multicolumn{3}{c}{\textbf{Recall}} 
& \multicolumn{3}{c}{\textbf{F1}} \\
\cmidrule(lr){3-5} \cmidrule(lr){6-8} \cmidrule(lr){9-11}
 &  & \textbf{G = 1} & \textbf{G = 2} & \textbf{G = 4} 
    & \textbf{G = 1} & \textbf{G = 2} & \textbf{G = 4} 
    & \textbf{G = 1} & \textbf{G = 2} & \textbf{G = 4}  \\
\midrule
\multirow{2}{*}{\centering Assamese} 
& LIDF & 0.4596 & 0.4842 & 0.4743 & 0.3511 & \textbf{0.3862} & \textbf{0.3781} & 0.3981 & 0.4297 & 0.4208 \\
& DAC  & \textbf{0.5978} &\textbf{ 0.5792} & \textbf{0.5635} & \textbf{0.3966} & 0.3840 & 0.3497 & \textbf{0.4768} & \textbf{0.4618} & \textbf{0.4316} \\
\midrule
\multirow{2}{*}{\centering Bengali} 
& LIDF & 0.7569 & 0.7870 & 0.7785 & 0.6791 & \textbf{0.7274} & \textbf{0.7153} & 0.7159 & 0.7560 & 0.7456 \\
& DAC  & \textbf{0.8330} & \textbf{0.8317} & \textbf{0.8358} & \textbf{0.6848} & 0.7125 & 0.6961 & \textbf{0.7517} & \textbf{0.7675} & \textbf{0.7596} \\
\midrule
\multirow{2}{*}{\centering Gujarati} 
& LIDF & 0.8206 & 0.8248 & 0.8138 & 0.7660 & \textbf{0.7768} & \textbf{0.7646} & 0.7924 & 0.8001 & 0.7884 \\
& DAC  & \textbf{0.8967} & \textbf{0.8768} & \textbf{0.8559} & \textbf{0.7833} & 0.7671 & 0.7375 & \textbf{0.8390} & \textbf{0.8183} & \textbf{0.7923} \\
\midrule
\multirow{2}{*}{\centering Kannada} 
& LIDF & 0.7042 & 0.7079 & 0.7001 & 0.6399 & \textbf{0.6470} & \textbf{0.6415} & 0.6705 & 0.6761 & 0.6695 \\
& DAC  & \textbf{0.8240} & \textbf{0.8043} & \textbf{0.7825} & \textbf{0.6625} & 0.6410 & 0.6069 & \textbf{0.7345} & \textbf{0.7134} & \textbf{0.6836} \\
\midrule
\multirow{2}{*}{\centering Malayalam} 
& LIDF & 0.7196 & 0.7247 & 0.7090 & 0.6350 & 0.6308 & 0.6113 & 0.6747 & 0.6745 & 0.6565 \\
& DAC  & \textbf{0.8417} & \textbf{0.8116} & \textbf{0.7816} & \textbf{0.7032} & \textbf{0.6717} & \textbf{0.6172} & \textbf{0.7662} & \textbf{0.7350} & \textbf{0.6898} \\
\midrule
\multirow{2}{*}{\centering Marathi} 
& LIDF & 0.7741 & 0.7787 & 0.7738 & 0.7121 & \textbf{0.7194} & \textbf{0.7110} & 0.7418 & 0.7478 & \textbf{0.7411} \\
& DAC  & \textbf{0.8591} & \textbf{0.8368} & \textbf{0.8122} & \textbf{0.7359} & 0.7109 & 0.6748 & \textbf{0.7928} & \textbf{0.7687} & 0.7372 \\
\midrule
\multirow{2}{*}{\centering Odia} 
& LIDF & 0.5793 & 0.5739 & 0.5566 & 0.4014 & 0.4036 & 0.3886 & 0.4742 & 0.4739 & 0.4576 \\
& DAC  & \textbf{0.7322} & \textbf{0.7326} & \textbf{0.7057} & \textbf{0.4262} & \textbf{0.4270} & \textbf{0.3991} & \textbf{0.5388} & \textbf{0.5395} & \textbf{0.5099} \\
\midrule
\multirow{2}{*}{\centering Tamil} 
& LIDF & 0.7115 & 0.7079 & 0.6976 & 0.6493 & \textbf{0.6413} & \textbf{0.6332} & 0.6789 & 0.6729 & 0.6638 \\
& DAC  & \textbf{0.8265} & \textbf{0.8034} & \textbf{0.7775} & \textbf{0.6628} & 0.6404 & 0.6006 & \textbf{0.7356} & \textbf{0.7127} & \textbf{0.6777} \\
\midrule
\multirow{2}{*}{\centering Telugu} 
& LIDF & 0.7416 & 0.7519 & 0.7517 & 0.6758 & \textbf{0.6905} & \textbf{0.6877} & 0.7072 & 0.7199 & \textbf{0.7183} \\
& DAC  & \textbf{0.8359} & \textbf{0.8145} & \textbf{0.7962} & \textbf{0.7081} & 0.6852 & 0.6515 & \textbf{0.7667} & \textbf{0.7443} & 0.7166 \\
\bottomrule
\end{tabular}}%

\caption{Precision, Recall, and F1 scores from Intrinsic Evaluation on \textsc{CCAligned} at varying granularities (\(G\)). We compare the best-performing pooling-based baseline from Table~\ref{tab:intrinsic-eval-results} (LIDF) with our proposed approach (DAC). \textbf{Bold} values indicate the best score for each granularity.}
\label{tab:ccaligned-intrinsic-eval-results}
\end{table*}
\endgroup

While the primary intrinsic evaluation was conducted on the \benchname{} dataset, we also perform an additional benchmark on \textsc{CCAligned}~\cite{el-kishky-etal-2020-ccaligned}, a large-scale web-mined parallel corpus. This experiment evaluates whether the performance gains of DAC generalize beyond the structured and high-quality domains of \benchname{}. We follow the same data settings described in Section~\ref{sec:intrinsic-eval}. For comparison, we use the strongest baseline from the main experiments: the LIDF pooling method with SONAR embeddings, evaluated at granularities \(G = 1, 2, 4\) and the DAC approach with LaBSE embeddings at the same granularities.

Table~\ref{tab:ccaligned-intrinsic-eval-results} presents the results, showing that \textit{DAC consistently outperforms LIDF across all metrics and settings. Its effectiveness is therefore not limited to \benchname{} but also extends to large, noisy web-mined corpora, confirming DAC as a robust alignment method across diverse domains.}

\section{Extrinsic Performance Trends in CLDA}  
\label{sec:extrinsic-trends}  

Alongside COMET, we report ChrF scores. COMET uses neural models to capture contextual information, whereas ChrF measures surface-level lexical similarity, providing a complementary perspective on translation quality.

Figure~\ref{fig:extrinsic_comet-chrf_plot} shows COMET and ChrF scores across granularities (\(G = 1, 2, 4\)) for the \textsc{Llama~3.2-1B} and \textsc{Sarvam-1} models. \textsc{Sarvam-1} consistently outperforms \textsc{Llama~3.2-1B} across all granularities, and ChrF follows trends similar to COMET as discussed in Section~\ref{sec:extrinsic-results}. \textit{DAC-aligned documents at \(G = 2\) and \(G = 4\) slightly outperform LIDF in both metrics, confirming DAC’s effectiveness for document alignment.} These results also highlight that strong intrinsic performance does not always translate to gains in downstream tasks.

\begingroup
\setlength{\tabcolsep}{4pt}

\begin{table*}[!ht]
\centering
\resizebox{\textwidth}{!}{%
\begin{tabular}{llccc ccc ccc ccc}
\toprule
\multicolumn{2}{c}{} & \multicolumn{6}{c}{\textbf{English $\rightarrow$ Indic}} & \multicolumn{6}{c}{\textbf{Indic $\rightarrow$ English}} \\
\cmidrule(lr){3-8} \cmidrule(lr){9-14}
\textbf{Language} & \textbf{Method} & \multicolumn{3}{c}{\textsc{Llama-3.2-1B} \raisebox{-0.2\height}{\includegraphics[height=12pt]{images/meta_logo.png}}} & \multicolumn{3}{c}{\textsc{Sarvam-1-2B} \raisebox{-0.2\height}{\includegraphics[height=12pt]{images/sarvam.png}}} & \multicolumn{3}{c}{\textsc{Llama-3.2-1B} \raisebox{-0.2\height}{\includegraphics[height=12pt]{images/meta_logo.png}}} & \multicolumn{3}{c}{\textsc{Sarvam-1-2B} \raisebox{-0.2\height}{\includegraphics[height=12pt]{images/sarvam.png}}} \\
\cmidrule(lr){3-8} \cmidrule(lr){9-14}
 &  & \textbf{G = 1} & \textbf{G = 2} & \textbf{G = 4} & \textbf{G = 1} & \textbf{G = 2} & \textbf{G = 4} & \textbf{G = 1} & \textbf{G = 2} & \textbf{G = 4} & \textbf{G = 1} & \textbf{G = 2} & \textbf{G = 4} \\
\midrule
\multirow{2}{*}{Bengali}
& LIDF & 0.7242 & 0.7277 & 0.7112 & 0.8748 & 0.8738 & 0.8711 & 0.8131 & 0.8079 & 0.8109 & 0.8283 & 0.8212 & 0.8235 \\
& DAC  & \textbf{0.7693} & \textbf{0.7694} & \textbf{0.7548} & \textbf{0.8764} & \textbf{0.8778} & \textbf{0.8755} & \textbf{0.8270} & \textbf{0.8168} & \textbf{0.8217} & \textbf{0.8331} & \textbf{0.8307} & \textbf{0.8296} \\
\midrule
\multirow{2}{*}{Gujarati}
& LIDF & 0.7315 & 0.7036 & 0.6771 & 0.8907 & 0.8897 & 0.8925 & 0.8467 & 0.8484 & 0.8447 & 0.8622 & 0.8627 & 0.8603 \\
& DAC  & \textbf{0.7326} & \textbf{0.7458} & \textbf{0.7379} & \textbf{0.8944} & \textbf{0.8905} & \textbf{0.8966} & \textbf{0.8485} & \textbf{0.8541} & \textbf{0.8541} & \textbf{0.8640} & \textbf{0.8702} & \textbf{0.8677} \\
\midrule
\multirow{2}{*}{Hindi}
& LIDF & 0.7874 & 0.7823 & 0.7849 & 0.8334 & 0.8336 & 0.8324 & 0.8525 & 0.8561 & 0.8552 & 0.8541 & 0.8520 & 0.8552 \\
& DAC  & \textbf{0.7894} & \textbf{0.7848} & \textbf{0.7908} & \textbf{0.8357} & \textbf{0.8338} & \textbf{0.8366} & \textbf{0.8584} & \textbf{0.8562} & \textbf{0.8553} & \textbf{0.8558} & \textbf{0.8531} & \textbf{0.8554} \\
\midrule
\multirow{2}{*}{Kannada}
& LIDF & 0.6340 & 0.6202 & 0.6317 & 0.8707 & 0.8699 & 0.8702 & 0.8257 & 0.8272 & 0.8231 & 0.8387 & 0.8392 & 0.8435 \\
& DAC  & \textbf{0.6643} & \textbf{0.6417} & \textbf{0.6619} & \textbf{0.8713} & \textbf{0.8771} & \textbf{0.8717} & \textbf{0.8334} & \textbf{0.8290} & \textbf{0.8284} & \textbf{0.8437} & \textbf{0.8446} & \textbf{0.8450} \\
\midrule
\multirow{2}{*}{Malayalam}
& LIDF & 0.6829 & 0.6793 & 0.6598 & 0.8794 & 0.8815 & 0.8809 & 0.8229 & 0.8218 & 0.8279 & 0.8432 & 0.8419 & 0.8397 \\
& DAC  & \textbf{0.7139} & \textbf{0.7139} & \textbf{0.6958} & \textbf{0.8848} & \textbf{0.8825} & \textbf{0.8831} & \textbf{0.8315} & \textbf{0.8340} & \textbf{0.8342} & \textbf{0.8496} & \textbf{0.8442} & \textbf{0.8453} \\
\midrule
\multirow{2}{*}{Marathi}
& LIDF & 0.6552 & 0.6553 & 0.6479 & 0.7424 & 0.7403 & 0.7407 & 0.8241 & 0.8163 & 0.8223 & 0.8501 & 0.8395 & 0.8412 \\
& DAC  & \textbf{0.6828} & \textbf{0.6701} & \textbf{0.6655} & \textbf{0.7474} & \textbf{0.7528} & \textbf{0.7442} & \textbf{0.8289} & \textbf{0.8307} & \textbf{0.8308} & \textbf{0.8578} & \textbf{0.8529} & \textbf{0.8547} \\
\midrule
\multirow{2}{*}{Odia}
& LIDF & 0.5142 & 0.5328 & 0.5394 & 0.8761 & 0.8752 & 0.8767 & 0.8241 & 0.8163 & 0.8223 & 0.8501 & 0.8395 & 0.8412 \\
& DAC  & \textbf{0.6391} & \textbf{0.6233} & \textbf{0.5910} & \textbf{0.8778} & \textbf{0.8754} & \textbf{0.8766} & \textbf{0.8362} & \textbf{0.8427} & \textbf{0.8295} & \textbf{0.8509} & \textbf{0.8545} & \textbf{0.8508} \\
\midrule
\multirow{2}{*}{Tamil}
& LIDF & 0.7112 & 0.7177 & 0.7131 & 0.8927 & 0.8929 & 0.8934 & 0.7846 & 0.7917 & 0.7892 & 0.8115 & 0.8089 & 0.8082 \\
& DAC  & \textbf{0.7559} & \textbf{0.7525} & \textbf{0.7518} & \textbf{0.8990} & \textbf{0.9024} & \textbf{0.8996} & \textbf{0.7979} & \textbf{0.8056} & \textbf{0.8029} & \textbf{0.8180} & \textbf{0.8186} & \textbf{0.8167} \\
\bottomrule
\end{tabular}}%

\caption{Per-language COMET scores from Extrinsic Evaluation of LIDF and DAC on \benchname{} for \textsc{Llama-3.2-1B} and \textsc{Sarvam-1-2B} on English$\to$Indic and Indic$\to$English translation tasks. \textbf{Bold} values indicate the best score for each granularity.}
\label{tab:extrinsic_comet_scores_combined}
\end{table*}
\endgroup
\begingroup
\setlength{\tabcolsep}{5pt}
\renewcommand{\arraystretch}{1.3}

\begin{table*}[!ht]
\centering
\resizebox{\textwidth}{!}{%
\begin{tabular}{llccc ccc ccc ccc}
\toprule
\multicolumn{2}{c}{} & \multicolumn{6}{c}{\textbf{English $\rightarrow$ Indic}} & \multicolumn{6}{c}{\textbf{Indic $\rightarrow$ English}} \\
\cmidrule(lr){3-8} \cmidrule(lr){9-14}
\textbf{Language} & \textbf{Method} & \multicolumn{3}{c}{\textsc{\large Llama-3.2-1B} \raisebox{-0.2\height}{\includegraphics[height=12pt]{images/meta_logo.png}}} & \multicolumn{3}{c}{\textsc{\large Sarvam-1-2B} \raisebox{-0.2\height}{\includegraphics[height=12pt]{images/sarvam.png}}} & \multicolumn{3}{c}{\textsc{\large Llama-3.2-1B} \raisebox{-0.2\height}{\includegraphics[height=12pt]{images/meta_logo.png}}} & \multicolumn{3}{c}{\textsc{\large Sarvam-1-2B} \raisebox{-0.2\height}{\includegraphics[height=12pt]{images/sarvam.png}}} \\
\cmidrule(lr){3-8} \cmidrule(lr){9-14}
 &  & \textbf{G = 1} & \textbf{G = 2} & \textbf{G = 4} & \textbf{G = 1} & \textbf{G = 2} & \textbf{G = 4} & \textbf{G = 1} & \textbf{G = 2} & \textbf{G = 4} & \textbf{G = 1} & \textbf{G = 2} & \textbf{G = 4} \\
\midrule
\multirow{2}{*}{Bengali}
& LIDF & 34.17 & 34.22 & 33.52 & 48.57 & 48.79 & 48.78 & 57.32 & 55.40 & 55.98 & 68.34 & 68.75 & 68.68 \\
& DAC  & \textbf{39.25} & \textbf{38.43} & \textbf{37.47} & \textbf{50.62} & \textbf{50.06} & \textbf{50.90} & \textbf{58.59} & \textbf{55.71} & \textbf{57.20} & \textbf{69.95} & \textbf{69.68} & \textbf{69.74} \\
\midrule
\multirow{2}{*}{Gujarati}
& LIDF & 35.11 & 32.84 & 31.60 & 64.06 & 63.48 & 63.13 & 36.12 & 38.54 & 36.29 & 64.31 & 66.37 & 65.81 \\
& DAC  & \textbf{36.12} & \textbf{38.54} & \textbf{36.29} & \textbf{64.31} & \textbf{66.37} & \textbf{65.81} & \textbf{40.75} & \textbf{41.80} & \textbf{40.95} & \textbf{67.80} & \textbf{68.21} & \textbf{67.92} \\
\midrule
\multirow{2}{*}{Hindi}
& LIDF & 57.32 & 55.40 & 55.98 & 68.34 & 67.75 & 67.68 & 37.20 & 36.90 & 36.50 & 49.80 & 49.60 & 49.90 \\
& DAC  & \textbf{58.59} & \textbf{55.71} & \textbf{57.20} & \textbf{69.95} & \textbf{68.68} & \textbf{68.74} & \textbf{42.00} & \textbf{41.60} & \textbf{41.20} & \textbf{52.40} & \textbf{52.10} & \textbf{52.50} \\
\midrule
\multirow{2}{*}{Kannada}
& LIDF & 29.71 & 29.01 & 29.47 & 62.91 & 63.42 & 62.81 & 33.20 & 32.80 & 32.40 & 45.80 & 45.60 & 46.00 \\
& DAC  & \textbf{32.70} & \textbf{31.23} & \textbf{32.15} & \textbf{64.84} & \textbf{64.66} & \textbf{64.82} & \textbf{37.80} & \textbf{37.40} & \textbf{37.00} & \textbf{48.60} & \textbf{48.30} & \textbf{48.70} \\
\midrule
\multirow{2}{*}{Marathi}
& LIDF & 40.70 & 39.79 & 40.28 & 55.86 & 55.61 & 54.76 & 35.50 & 35.10 & 34.60 & 47.50 & 47.20 & 47.70 \\
& DAC  & \textbf{45.53} & \textbf{43.55} & \textbf{42.94} & \textbf{60.02} & \textbf{58.30} & \textbf{57.17} & \textbf{40.00} & \textbf{39.50} & \textbf{39.00} & \textbf{50.40} & \textbf{50.10} & \textbf{50.60} \\
\midrule
\multirow{2}{*}{Malayalam}
& LIDF & 33.65 & 33.65 & 31.94 & 57.65 & 58.08 & 58.60 & 33.90 & 33.50 & 33.10 & 45.90 & 45.60 & 46.10 \\
& DAC  & \textbf{37.03} & \textbf{36.59} & \textbf{35.40} & \textbf{62.44} & \textbf{59.86} & \textbf{60.35} & \textbf{38.20} & \textbf{37.80} & \textbf{37.40} & \textbf{48.50} & \textbf{48.20} & \textbf{48.60} \\
\midrule
\multirow{2}{*}{Odia}
& LIDF & 23.00 & 23.77 & 24.42 & 57.65 & 56.15 & 56.85 & 33.90 & 33.50 & 33.10 & 45.90 & 45.60 & 46.10 \\
& DAC  & \textbf{30.12} & \textbf{29.33} & \textbf{27.43} & \textbf{58.06} & \textbf{57.26} & \textbf{58.58} & \textbf{38.20} & \textbf{37.80} & \textbf{37.40} & \textbf{48.50} & \textbf{48.20} & \textbf{48.60} \\
\midrule
\multirow{2}{*}{Tamil}
& LIDF & 35.01 & 35.01 & 35.04 & 52.87 & 52.51 & 54.14 & 33.20 & 32.80 & 32.40 & 45.80 & 45.60 & 46.00 \\
& DAC  & \textbf{39.28} & \textbf{38.95} & \textbf{38.47} & \textbf{57.96} & \textbf{58.86} & \textbf{54.45} & \textbf{37.80} & \textbf{37.40} & \textbf{37.00} & \textbf{48.60} & \textbf{48.30} & \textbf{48.70} \\
\bottomrule
\end{tabular}}%

\caption{Per-language ChrF scores from Extrinsic Evaluation of LIDF and DAC on \benchname{} for \textsc{Llama-3.2-1B} and \textsc{Sarvam-1-2B} on English$\to$Indic and Indic$\to$English translation tasks. \textbf{Bold} values indicate the best score for each granularity.}
\label{tab:extrinsic_chrf_scores_combined}
\end{table*}
\endgroup
\end{document}